%% file: main.tex
\documentclass[10pt]{article} %

\newif\ifanon
\anonfalse
\newcommand{\anon}[2]{\ifanon#1\else#2\fi}

\ifanon
  \usepackage{tmlr}             %
\else
  \usepackage[preprint]{tmlr}
\fi

\input{math_commands.tex}

\usepackage[pagebackref=true,
    colorlinks=true,
    allcolors=blue!70!black
]{hyperref}

\usepackage{url}
\usepackage{graphicx}
\usepackage{amsmath}
\usepackage{amssymb}
\usepackage{bm}
\usepackage{booktabs}
\usepackage{multirow}
\usepackage{subcaption}
\usepackage{algorithm}
\usepackage{algpseudocode}

\graphicspath{{figures/}}

\newcommand{\What}{\widehat{W}}              %
\newcommand{\grow}{\gamma_{\text{row}}}      %
\newcommand{\gcol}{\gamma_{\text{col}}}      %
\newcommand{\dout}{d_{\text{out}}}
\newcommand{\din}{d_{\text{in}}}
\newcommand{\reduceop}{\mathrm{reduce}}
\newcommand{\OptStep}{\mathrm{OptStep}}
\newcommand{\AdamStep}{\mathrm{AdamStep}}
\DeclareMathOperator{\diag}{diag}

\title{Improving Neural Network Training by\\Decoupling the Magnitude and Direction of Weight Vectors}

\AddToHook{para/begin}{\looseness=-1}

\author{\name Alexander H\"agele$^{*}$ \email alexander.hagele@epfl.ch \\
      \addr Machine Learning and Optimization Lab, EPFL
      \AND
      \name Alejandro Hern\'andez-Cano$^{*}$ \email alejandro.hernandezcano@epfl.ch \\
      \addr Machine Learning and Optimization Lab, EPFL
      \AND
      \name Atli Kosson\footnotemark[2] \email atli.kosson@epfl.ch \\
      \addr Machine Learning and Optimization Lab, EPFL
      \AND
      \name Martin Jaggi\footnotemark[2] \email martin.jaggi@epfl.ch \\
      \addr Machine Learning and Optimization Lab, EPFL}

\def\month{06}  %
\def\year{2026} %
\def\openreview{\url{https://openreview.net/forum?id=XXXX}} %

\makeatletter
\def\section{\@startsection{section}{1}{\z@}{-2.0ex plus -0.5ex minus -.2ex}{1.0ex plus 0.2ex minus 0.1ex}{\large\bf\raggedright\sffamily}}
\def\subsection{\@startsection{subsection}{2}{\z@}{-1.6ex plus -0.5ex minus -.2ex}{0.5ex plus .1ex}{\normalsize\bf\raggedright\sffamily}}
\def\subsubsection{\@startsection{subsubsection}{3}{\z@}{-1.3ex plus -0.5ex minus -.2ex}{0.3ex plus .1ex}{\normalsize\bf\raggedright\sffamily}}
\makeatother

\begin{document}

\maketitle
\ifanon\else
\begingroup
\renewcommand{\thefootnote}{\fnsymbol{footnote}}
\footnotetext[1]{Equal Contribution. Correspondence to \texttt{alexander.hagele@epfl.ch}.}
\footnotetext[2]{Equal Senior Contribution.}
\endgroup
\fi

\begin{abstract}
Modern neural network training relies on optimizers such as Adam and Muon which act on each weight matrix as a single object.
Yet every weight matrix carries two distinct quantities --- a \emph{magnitude} and a \emph{direction} --- and all optimizers stepping in the matrix as a whole couple their dynamics: the directional change from an update depends on the current magnitude, while the magnitude drifts as a byproduct of learning the direction. Then, neither is directly governed by the learning rate. Typical training therefore leans on surrounding recipes such as weight decay and warmup to keep learning stable at scale, though these regulate the coupling only indirectly. Other recent methods instead constrain the weight to a fixed-norm sphere, but add no learnable magnitude, leaving scale control to normalization layers alone.
We propose \emph{Magnitude--Direction (MD) Decoupling}, an optimizer modification that factorizes each weight into a fixed-norm direction on a hypersphere and learnable per-row and per-column magnitude gains, updated at separate learning rates, all while the model still sees a single fused weight tensor. The method is agnostic to the base optimizer and removes the need for weight decay and warmup. Across both Adam and Muon, MD Decoupling improves on well-tuned baselines, transfers the optimal LR across model width without retuning, and continues to help at scale on large Mixture-of-Experts (MoE) models. Treating magnitude and direction as separately controlled quantities thus yields more predictable training dynamics and a simple, broadly applicable improvement to modern optimizers.\anon{}{\footnote{{A shorter and more accessible version of this paper is accessible at \url{https://haeggee.github.io/posts/magnitude-direction-decoupling}.}}}
\end{abstract}

\section{Introduction}

Much recent progress in neural network training comes from rethinking what a good update to a weight \emph{matrix} should be. Beyond Adam~\citep{kingma2014adam}, matrix-aware optimizers such as Shampoo~\citep{gupta2018shampoo}, SOAP~\citep{vyas2024soap}, and Muon~\citep{jordan2024muon} improve the update by accounting for the geometry of the weight space~\citep{bernstein2024norm,pethick2025scion}. At scale, making training balanced and predictable further relies on an apparatus of intricate rules and recipes
~\citep{everett2024scaling,wang2024adamw,dey2025completep,mlodozeniec2025completed,dial2026_pretraining_speedup}.

\ifanon
\begin{figure}[t]
\else
\begin{figure}[t]
\fi
      \centering
      \begin{subfigure}{0.325\textwidth}\includegraphics[width=\textwidth]{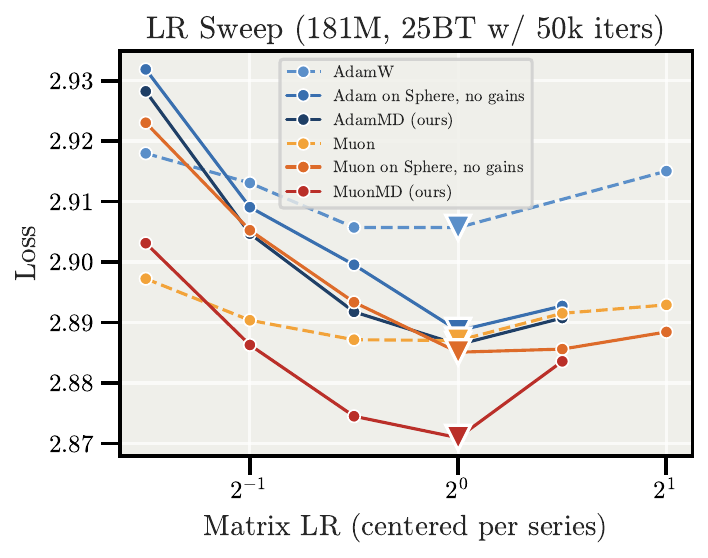}\end{subfigure}\hfill
      \begin{subfigure}{0.325\textwidth}\includegraphics[width=\textwidth]{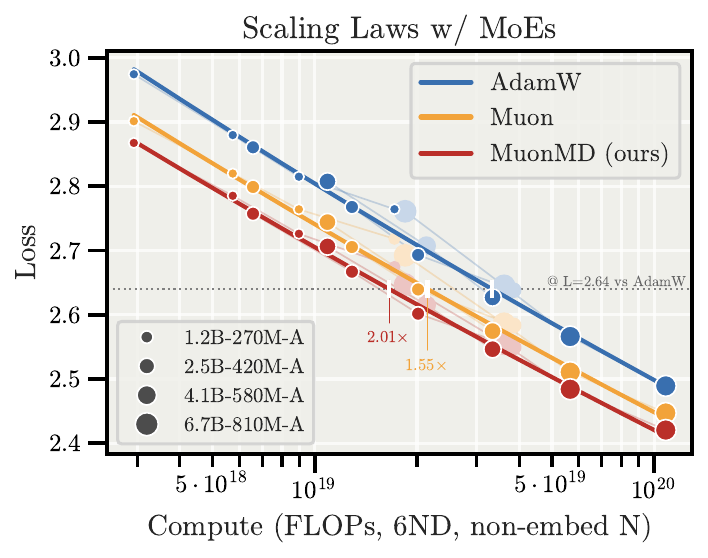}\end{subfigure}\hfill
      \begin{subfigure}{0.325\textwidth}\includegraphics[width=\textwidth]{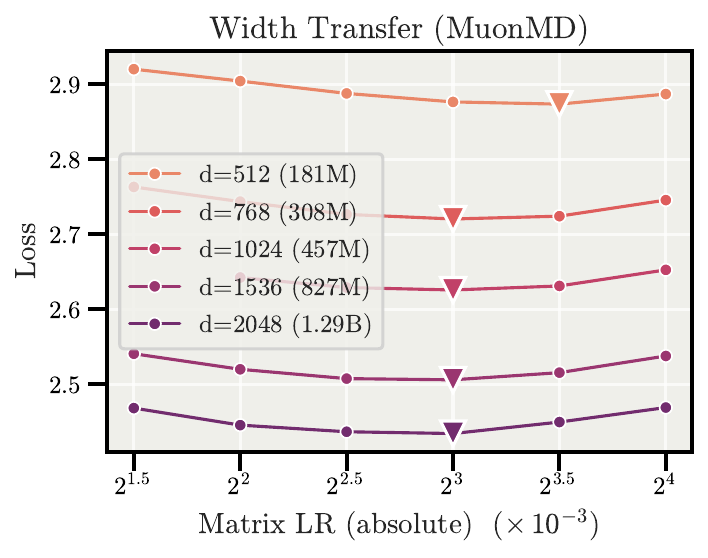}\end{subfigure}
      \caption{\textbf{Magnitude--Direction (MD) Decoupling improves on well-tuned Adam and Muon, keeps the improvement across compute on large MoEs, and makes the optimal learning rate transfer across model width.} Three views of the method (full details in \autoref{sec:details}). \figleft~Learning-rate sweep on a dense model: independently of the base optimizer, fixing the weights onto a sphere improves the optimal loss, and adding learnable magnitudes (our MD variant) gives a further boost. \figcenter~Scaling laws for sparse MoEs, where the improvement holds across a wide range of compute. \figright~LR transfer across model width: controlling the relative weight update directly through the sphere makes the optimal LR transfer without retuning.}
      \label{fig:headline}
\end{figure}

However, all matrix optimizers still share the same underlying mechanics of neural networks: a network is a stack of many weight matrices, and every matrix $W$ carries two distinct quantities, a magnitude $\lVert W\rVert$ and a direction $\What = W/\lVert W\rVert$, like a vector in polar coordinates. Most of learning is \emph{rotation} of that direction~\citep{wan2021spherical}. Stepping in $W$ as a whole couples the two: the angular change scales inversely with the current magnitude, while the magnitude itself drifts upward as a byproduct of learning the direction, so the learning rate controls neither (\autoref{sec:problem}). As a result, standard optimizers struggle to learn the magnitude of weight matrices and lean on weight decay to keep learning the direction over the long term~\citep{kosson2023rotational}. Warmup~\citep{goyal2017accurate,xiong2020layer} and fixes like QK-clip~\citep{kimiteam2025kimi} patch similar symptoms of runaway growth.

\textbf{Magnitude--Direction Decoupling.} 
We therefore propose \emph{Magnitude--Direction (MD) Decoupling}, an optimizer modification that factorizes each weight into a fixed-norm direction on a hypersphere and learnable per-row and per-column magnitude gains, updating the two at separate learning rates (\autoref{sec:solution}). The factorization echoes classic Weight Normalization~\citep{salimans2016weight}, but puts the direction on a \emph{fixed} sphere with a normalized update and learns the gains at their own rate: the learning rate then sets the angular update directly, while the gains recover the fine-grained scale control that fixing the norm gives up. The split lives entirely inside the optimizer (\autoref{sec:solution}): the model sees a single fused weight tensor, the method is agnostic to the base optimizer, and it effectively removes the need for weight decay and warmup.

\textbf{Context in the literature.} 
Our work is not isolated. In fact, many works, both longstanding~\citep{you2017large,liu2018decoupled,liu2021learning,karras2023analyzing} and concurrent works, reach related ideas. One recent line fixes the weights to a sphere and drops weight decay~\citep{loshchilov2024ngpt,muonh2026,franke2025learning,ren2026rethinking,bernstein2025modular}, but adds no learnable magnitude, leaving scale control to normalization-layer gains. Others learn explicit scales without a sphere constraint~\citep{velikanov2026learnable,wang2026negligible}, or split each weight into a per-row magnitude and direction under Muon~\citep{lion2026muown,huebler2026angularmuown}. Developed largely in parallel, these threads often differ only in a few choices: which norm is fixed and along which axis, which optimizer is used, or whether a separate magnitude is added. A central motivation of our work is to unify them through controlled experiments and discussion in related work (\autoref{sec:related} \& \autoref{sec:appendix-relwork}), asking which choices matter in practice for training large language models.

\textbf{Findings.} We find that across both Adam and Muon, MD Decoupling improves on well-tuned baselines and transfers the optimal learning rate across width without retuning, in the spirit of $\mu$P~\citep{yang2022tensor} but obtained directly from the sphere~\citep{kosson2025weight,muonh2026,ren2026rethinking}. It keeps its edge at scale on large Mixture-of-Experts models~\citep{shazeer2017outrageously,dai2024deepseekmoe}, reaching AdamW's loss with ${\sim}2\times$ less compute (\autoref{fig:headline}). Our controlled ablations on dense models (\autoref{sec:details}) let us settle on a default recipe, and additionally cover the gain parametrization, the schedule, depth scaling, warmup-free and continual training, and a comparison to nGPT~\citep{loshchilov2024ngpt}. We discuss implications and research questions that require further understanding in \autoref{sec:discussion}.

\section{Magnitude--Direction Interference}
\label{sec:problem}
\looseness-1
We start by revisiting how the magnitude and direction of the weight interfere in a toy example in \autoref{fig:interference}. Here, the loss is \emph{scale-invariant}: This means that only the direction of the weights affects the output, not the magnitude. This is a common case in deep learning, where matrices are often followed by normalization layers~\citep{ba2016layer,zhang2019rmsnorm}. Yet, the magnitude shapes what a single optimizer step does, through two effects that the learning rate fails to control; we look at each effect in turn. Together they are why \emph{standard optimizers struggle to learn the magnitude of weight matrices and lean on weight decay to keep learning the direction over the long term}~\citep{kosson2023rotational}. 
Throughout, unless otherwise noted, $\lVert\cdot\rVert$ is the Frobenius norm, and ``direction'' / ``magnitude'' may refer to the whole matrix or to its rows and columns depending on the variant.

\begin{figure}[t]
\centering
\includegraphics[width=\textwidth]{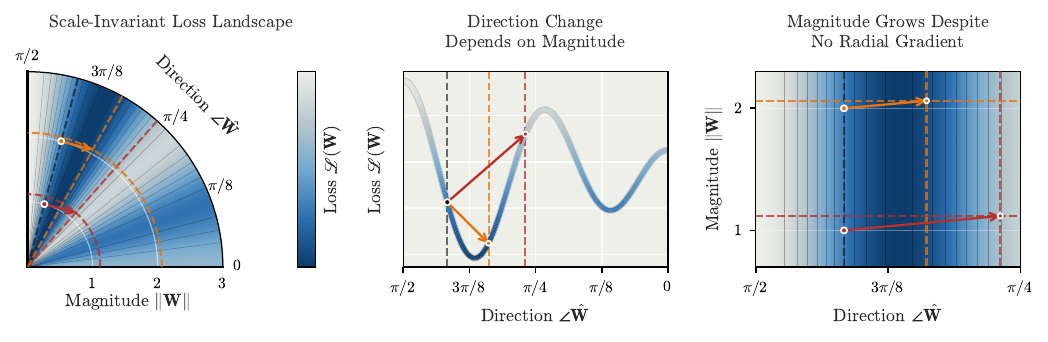}
\caption{\textbf{In standard optimizers the weight magnitude silently distorts each update: the same step rotates the weights more at small magnitude and inflates the norm even when only the direction matters.} Illustrated on a toy scale-invariant loss, where only the direction of the weights affects the loss. \figleft~The loss landscape in polar coordinates, with the same normalized optimizer step taken from a small (red) and a large (orange) starting magnitude. \figcenter~The identical step changes the direction --- and hence the loss --- far more at small magnitude than at large magnitude. \figright~Even though the loss has no radial gradient, the step still increases the magnitude.}
\label{fig:interference}
\end{figure}

\textbf{Direction change depends on magnitude.} We can measure the directional change caused by an optimizer update $\Delta W$ through the angular update $\angle(W, W+\Delta W)$~\citep{wan2021spherical}, which is closely approximated by the relative update $\lVert\Delta W\rVert/\lVert W\rVert$. For a normalized optimizer like Adam or Muon the update size is set by the LR and is independent of the weight norm, so the directional change is roughly inversely proportional to the current magnitude $\lVert W\rVert$ (middle panel of \autoref{fig:interference}). The LR therefore does not directly set the rate of directional change, and that rate can vary across layers and over time in ways that hurt learning. Prior work on Rotational Equilibrium~\citep{kosson2023rotational} showed how weight decay partially fixes this by modulating the relative updates over time and balancing them across layers.

\textbf{Magnitude grows despite no radial gradient.} Direction changes also feed back into the magnitude. In practice, updates tend to be roughly perpendicular to the current weights --- from properties of scale-invariance or from noise --- and a perpendicular update \emph{always} increases the magnitude~\citep{heo2021adamp}. This happens even when nothing pulls the weights outward: a scale-invariant function has no radial gradient, yet the norm still creeps up (\autoref{fig:interference}, \panelright). For a non-scale-invariant function the (negative) radial signal has to be strong enough to counteract it. In practice the magnitudes converge toward an equilibrium set by the LR and weight decay rather than any learned optimum~\citep{kosson2023rotational}, and this unnecessary growth can require tricks like Kimi's QK-clip~\citep{kimiteam2025kimi} to tame.

We describe the magnitude as a single scalar for simplicity, but the same interference applies per row or per column: an optimizer that cannot learn the per-matrix scale well will not do better at finer granularity.

\section{Magnitude--Direction Decoupling}
\label{sec:solution}

The fix to the interference between weight norms and updates is to optimize the weights in a form that \emph{resembles} polar coordinates --- a direction and a magnitude --- updating each separately so neither interferes with the other. Concretely, we factorize each weight into a direction $\What$ with a fixed norm (so it lies on a fixed hypersphere) and learnable magnitude gains:
\begin{equation}
W = \diag(\grow)\,\What\,\diag(\gcol), \qquad \What\ \text{on the sphere},
\end{equation}
with $\grow\in\R^{\dout}$ and $\gcol\in\R^{\din}$ learnable gains (a single scalar or a one-sided gain are special cases). The two are learned at separately controlled rates; the update rules are in \autoref{sec:optimizer}.

\textbf{Fine-grained scales.} Our previous section argued how normalization layers make the loss invariant to a \emph{single overall scalar} per matrix. However, this is not the case for finer-grained scales. The model still needs to control the scale of its activations, amplifying some features while damping others, and mixing activations that live at different scales. Per-row and per-column magnitudes change the function, and being able to learn them matters. In fact, this is why the learnable gains in RMSNorm layers help. But a standard transformer has far fewer such gains than its matrices have rows and columns, and normalization layers are not everywhere, so existing gains cannot provide the same fine-grained control. Our gains $\grow, \gcol$ make this control explicit and learned at a well-regulated speed. In theory, only $\gcol$ gains can be redundant, and only if a normalization layer with gains precedes the matrix.

\subsection{The Decoupled Optimizer Step}
\label{sec:optimizer}

\textbf{Updating the direction.} We keep the size of the update to $\What$ proportional to its magnitude, then project $\What$ back onto the sphere so the magnitude stays constant. The relative weight update is then determined by the LR at every step (for optimizers that produce normalize updates, as is done in practice). With no equilibrium to drift toward and no dependence on the initialization norm or training length, the LR schedule directly sets the relative update.

\textbf{Updating the magnitude.} The magnitude of $W$ is determined by the gains $\gamma$, which are updated like other learnable gains typically found in normalization layers. The gains can be either a scalar, a vector acting on each row or column, or two vectors scaling both the rows and columns. We note that these magnitude gains do not provide any additional representational capacity over the original matrix; they only affect the learning dynamics.

\textbf{Fused weights.} In practice, we do not want to keep $\gamma$ and $\What$ separate and reconstruct the weights in the forward and backward pass. This adds unnecessary round trips through memory. Instead, the model holds the \emph{fused weight tensor} $W$ and computes the gradient $G = \partial L/\partial W$ as usual. Then, at each step, the optimizer recovers the direction and the gain, splits the gradient between them, updates each, projects the direction back onto the sphere, and reassembles $W$.

\autoref{alg:md-step} illustrates the optimizer step in the simplest case, a single scalar $\gamma$ with $W=\gamma\odot \What$. The general version with per-row and per-column gains is given in \autoref{alg:md-step-full} in \autoref{sec:full-algorithm}.

\begin{algorithm}[t]
\caption{Magnitude--Direction decoupled optimizer step (scalar gain $\gamma$, fused weight $W=\gamma\odot\What$).}
\label{alg:md-step}
\begin{algorithmic}[1]
\Require fused weight $W$, gain $\gamma$, gradient $G=\partial L/\partial W$, direction LR $\eta_W$, gain LR $\eta_\gamma$, sphere radius $c_F$.
\State $\What \leftarrow W / \gamma$ \Comment{recover the on-sphere direction}
\State $g_\gamma \leftarrow \reduceop\big(\What \odot G\big)$ \Comment{gain gradient: sum over the axis the gain does not span}
\State $G_{\What} \leftarrow \gamma \odot G$ \Comment{direction gradient $\partial L/\partial \What$}
\State $\What \leftarrow \OptStep\big(\What,\, G_{\What},\, \eta_W\big)$ \Comment{any (normalized) matrix optimizer (Adam / Muon / \dots)}
\State $\What \leftarrow c_F \cdot \, \What \,/\, \lVert \What\rVert$ \Comment{project back onto the sphere of fixed radius $c_F$}
\State $\gamma \leftarrow \AdamStep\big(\gamma,\, g_\gamma,\, \eta_\gamma\big)$ \Comment{step the gain (its own LR)}
\State $W \leftarrow \gamma \odot \What$ \Comment{reassemble for the next forward}
\end{algorithmic}
\end{algorithm}

\textbf{Reparameterized gain.} The gain can be updated in several ways --- directly, kept strictly positive, or learned at a controlled pace --- e.g.\ by storing a ``raw'' gain $\widehat{\gamma}$ and applying a positive map $\gamma = \varphi(\widehat{\gamma})$ such as softplus. We ablate these choices in the results below (\autoref{sec:gains}) and find only a minimal edge for softplus, which we adopt as the default; the optimal parametrization might be more complex.

\looseness-1
\textbf{Important Properties.} We explicitly restate the benefits of separating magnitude and direction.
\begin{itemize}
\item The core idea behind Magnitude-Direction Decoupling is \emph{independent of the optimizer}: the weight update can be treated as a black box, so it naturally fits different optimizers (AdEMAMix, Muon, Shampoo, \dots).
\item \emph{We no longer need weight decay}, since the weights are already on the sphere. This also avoids its complicated interactions with the LR schedule, and the effective step size is now just the LR.
\item \emph{We get LR transfer} in width for any sufficiently long training run, because we control the relative weight update directly.\footnote{The exact transfer may be mildly optimizer-dependent; e.g., for Muon, it may rely on matching the update RMS to that of the weight norm sphere.}
\item Like Muon, \emph{we no longer need warmup}, since the large early updates it exists to prevent never appear. In fact, in our experiments, we see this strongly improves the final loss, since the early stages of training are spent with higher effective learning.\footnote{For extremely large models, we believe Adam might still need a short warmup to prevent early instability due to cold momentum states.}
\end{itemize}

\section{Empirical Evaluation}
\label{sec:details}

This section is organized in two parts. In \autoref{sec:dense}, we use small dense models to ablate the components of the method --- the normalization and gain axes, the embedding normalization, learning-rate transfer across width and depth, the learning-rate schedule, and warmup-free and continual training --- and settle on a default recipe. In \autoref{sec:moe}, we then verify that this recipe holds at scale, training large Mixture-of-Experts models and comparing against well-tuned baselines.

\subsection{Ablations on Dense Models}
\label{sec:dense}

\textbf{Setup.} For the dense ablations we use GPT-style language models from 181M to 1.29B parameters, each with head-dimension 128, GQA~\citep{ainslie2023gqa}, QK-norm~\citep{dehghani2023scaling,wortsman2023small}, and Sandwich Norm~\citep{ding2021cogview,kim2025peri} with RMSNorm~\citep{zhang2019rmsnorm}. We apply a fixed scale $\alpha=\frac{1}{L}$ to the block-output after the RMSNorm (for proper depth scaling; more in \autoref{sec:transfer}). Matrix parameters are initialized with standard deviation $\frac{1}{\sqrt{d}}$, and the embeddings are upscaled by $\sqrt{d}$ to give an RMS of $1$ going into the model. 
We give the full architecture and hyperparameter details in \autoref{sec:appendix-setup}.

\textbf{Base comparison.} Our ablation base is the 181M model ($d=512, L=12$) trained on 25B tokens of a FineWeb-Edu~\citep{penedo2024fineweb} subset. This is deliberate \emph{strong overtraining} (Chinchilla sense): at $50$k steps with a batch size of ${\sim}0.5$M tokens (4096 sequence length),
At this scale, longer-term training dynamics become visible, is a closer match to real pretraining runs with potentially millions of steps.

\textbf{Sweep setups.} We focus on AdamW and Muon as the most popular base optimizers. Across all experiments (incl. the \autoref{fig:headline} sweeps), we \emph{fix} the LR of every Adam-optimized parameter group shared across optimizers (e.g. embeddings, RMSNorm gains, etc) at values we verified to be in a good range (see \autoref{fig:lr-panels-combined} in \autoref{sec:appendix-lr-sweeps}); $10^{-3}$ for the output layer, $3\cdot10^{-3}$ for embeddings) and sweep the \emph{matrix LR} separately for each optimizer or setup change. This means every method is tuned with the \emph{same budget}. The standard methods (AdamW and Muon) use weight decay $0.1$; the MD variants use none, since the weights are already on the sphere. For Muon in \autoref{fig:headline} we use a scale factor of $\sqrt{\frac{\dout}{\din}}$ (which we found to be noticeably better than the RMS grafting when sweeping (see \autoref{fig:muon-scale-modes}). Unless noted otherwise, dense models use a linear LR decay to $10^{-8}$ for all groups. For the ablations below, we default to Muon as the matrix optimizer under MD decoupling.

\subsubsection{Normalization Axis}
\label{sec:axes}

\begin{figure}[t]
\centering
\begin{subfigure}{0.4\textwidth}\centering\includegraphics[width=0.92\textwidth]{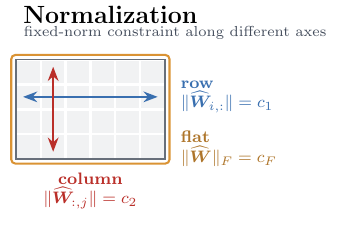}\end{subfigure}\quad
\begin{subfigure}{0.4\textwidth}\centering\includegraphics[width=0.92\textwidth]{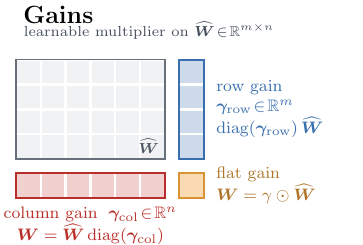}\end{subfigure}
\caption{\textbf{Magnitude--Direction Decoupling has two independent per-matrix choices: which axis the direction is normalized along, and which axis the learnable gain acts on.} \figleft~The axis along which a matrix can be constrained, and \figright~The axis along which the gains can act: row, column, both, or flat / Frobenius.}
\label{fig:axes}
\end{figure}

There are two independent choices per matrix (\autoref{fig:axes}): along which axis we constrain the direction (each output row to unit norm, each input column, or the whole matrix in Frobenius norm), and along which axis the gain is free to act. We take these in turn below: the normalization axis, then the special case of the embeddings, and finally the magnitude gains.

\textbf{Normalization axis.} A priori it is unclear which constraint should work best: EDM2~\citep{karras2023analyzing} keeps each row (output channel) at a fixed norm, nGPT~\citep{loshchilov2024ngpt} alternates rows for QKV + up projections and columns for down projections, and AdamH/MuonH~\citep{muonh2026} use the Frobenius norm. In each case, we hold the direction at its initialization norm so the constraint does not change the model at the start of training; with our $\frac{1}{\sqrt{d}}$ initialization this is a Frobenius target of $\sqrt{\max(\dout,\din)}$ for a matrix $\What\in \R^{\dout \times \din}$. We keep embeddings and the output (LM-head) rows at unit $L_2$ norm throughout, independent of the sphere mode.

\textbf{Update step scale.} We do not renormalize the update from the base optimizer. For Muon, we rescale its output by a fixed $\sqrt{\max\!\big(\frac{\dout}{\din},\, \frac{\din}{\dout}\big)}$, which makes the update RMS (assuming proper orthogonalization through Newton--Schulz) match the RMS of the weight norm under our $\frac{1}{\sqrt{d}}$ initialization. The factor is therefore set by the target weight norm and should be adapted whenever that norm changes --- for instance under a scaled output-projection initialization, as in our MoE runs (\autoref{sec:appendix-setup-moe}). We show sweeps for the available conventions in \autoref{fig:muon-scale-modes}, \autoref{sec:appendix-lr-sweeps}. We also ablated removing the radial component of the direction's gradient (projecting it onto the tangent space of the sphere), which made no measurable difference.

\textbf{Results.} Across the three constraints, the final losses turn out to be nearly identical at their optimum (\autoref{fig:norm-axis}). Note that we do not use gains for MD in this comparison. We therefore adopt the Frobenius constraint, since it is the most flexible: it only fixes the overall scale of the matrix and leaves the relative scale of its rows and columns open.

\textbf{Comparison to nGPT.} Our fixed-norm motivation is closely related to nGPT~\citep{loshchilov2024ngpt}, which also trains on a sphere without weight decay but bundles this with several architectural changes ($L_2$ normalization, an interpolated residual, and a reduced base scale on its learnable vectors). Disentangling the optimizer and architecture in \autoref{sec:appendix-ngpt} (\autoref{fig:ngpt}), we find that as proposed nGPT beats our base architecture, but our magnitude--direction decoupling \emph{on top of} nGPT's architecture surpasses nGPT itself, with an even larger margin under Muon.

\subsubsection{Embeddings}
\label{sec:embed-norm}

\textbf{Per-row unit norm.} The embeddings of the network are a special case. Each row is a single token's vector that the model looks up independently, so the natural constraint is per-row --- unit $L_2$ norm on each embedding --- rather than the per-matrix Frobenius sphere we use for the other weights. To understand the impact of normalization on embeddings, we ablate this part of the network separately in \autoref{fig:embed-norm}: constraining every embedding vector to unit norm versus leaving its norm free to vary. The left panel sweeps the matrix LR for each mode, while the center and right panels track the loss and the relative embedding update over training. Keeping each embedding at unit norm performs at least on par with leaving it free while keeping the embedding update better-behaved, so we adopt it as our default and hold embeddings (and the LM-head rows) at unit $L_2$ norm throughout, independent of the matrix sphere mode.

\textbf{Relation to post-embedding RMSNorm.} This unit-norm constraint is closely related to a now-common trick of placing an RMSNorm directly after the embedding layer, as used in the nanoGPT speedruns~\citep{jordan2024nanogpt}: normalizing each token's embedding vector is exactly the per-row constraint we apply, only enforced in the optimizer on the embedding weights themselves rather than as a module in the forward pass (no extra forward/backward computation and no architectural change). 

\begin{figure}[t]
\centering
\begin{subfigure}{0.325\textwidth}\centering\includegraphics[width=\textwidth]{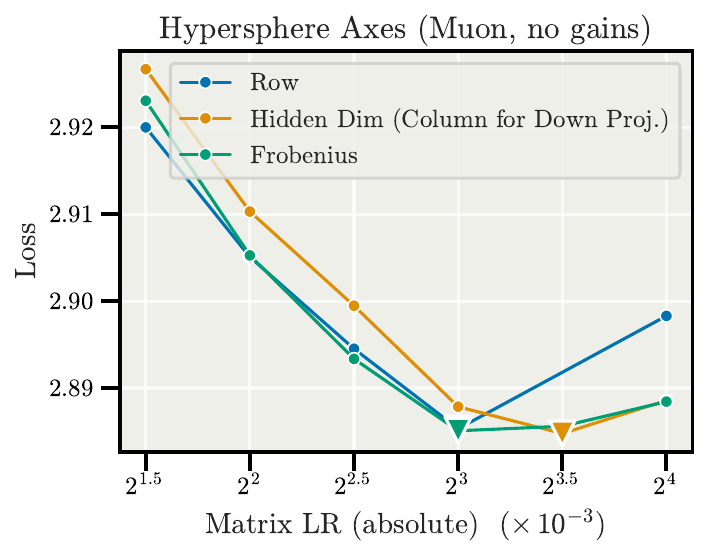}\end{subfigure}\hspace{0.03\textwidth}
\begin{subfigure}{0.325\textwidth}\centering\includegraphics[width=\textwidth]{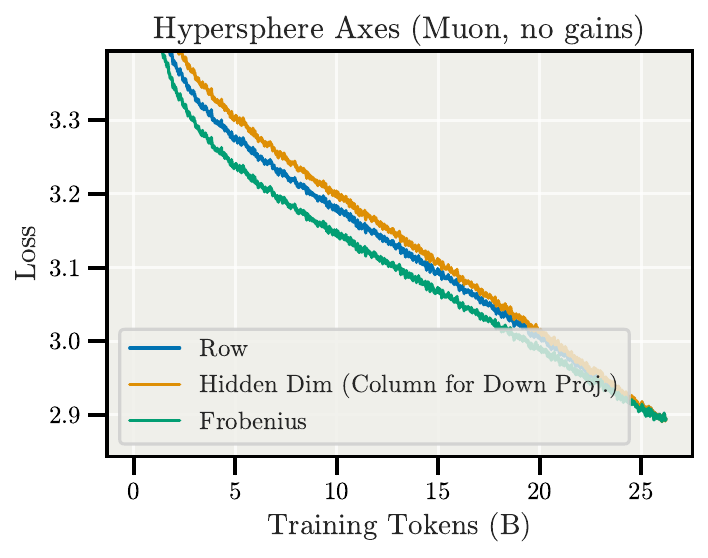}\end{subfigure}
\caption{\textbf{The choice of normalization axis barely affects the final loss, so we adopt the most flexible Frobenius constraint.} Comparison of constraining each output row, each input column, or the whole matrix (Frobenius) to a fixed norm, on the 181M dense model (25B tokens), without gains. \figleft~LR sweep of the final loss for each normalization mode. \figright~The corresponding loss curves over training.}
\label{fig:norm-axis}

\medskip

      \begin{subfigure}{0.325\textwidth}\centering\includegraphics[width=\textwidth]{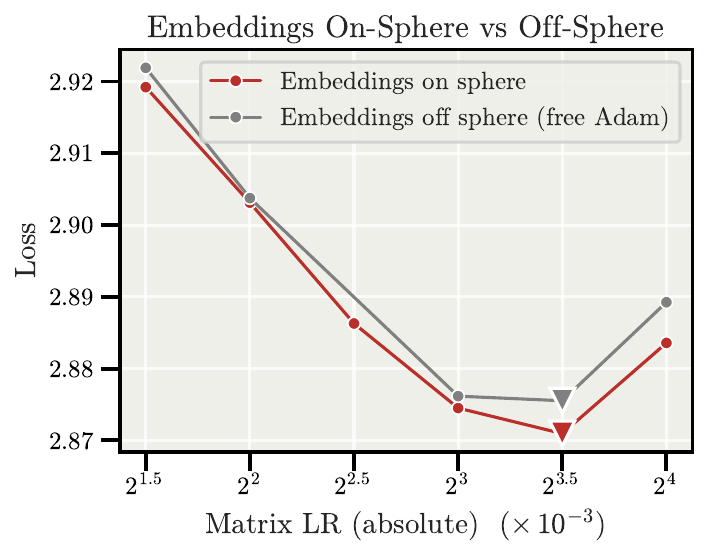}\end{subfigure}\hfill
      \begin{subfigure}{0.325\textwidth}\centering\includegraphics[width=\textwidth]{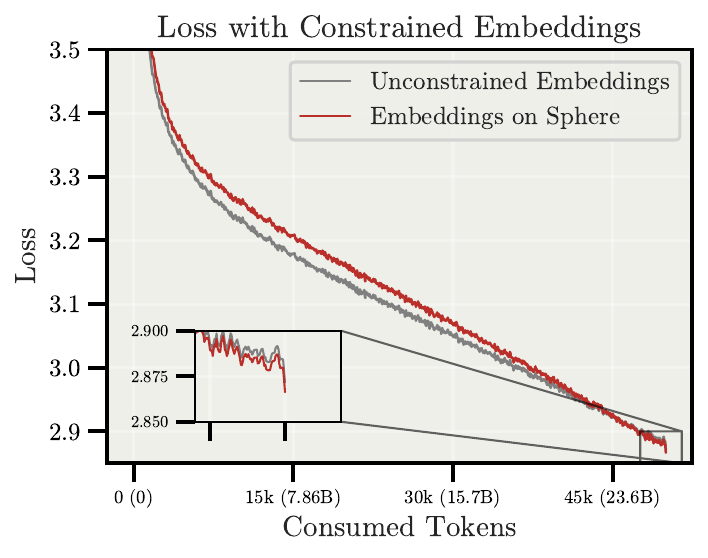}\end{subfigure}
      \begin{subfigure}{0.325\textwidth}\centering\includegraphics[width=\textwidth]{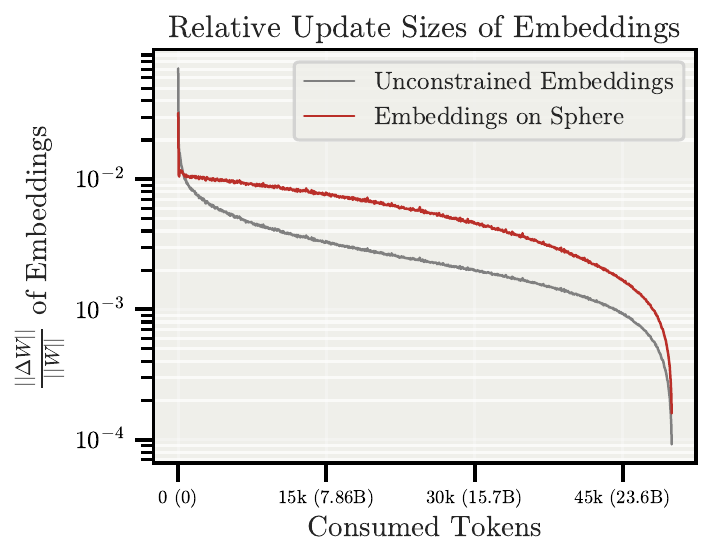}\end{subfigure}
      \caption{\textbf{Holding each embedding vector at unit norm performs slightly better than leaving them unconstrained, while keeping the embedding update better behaved.} Ablation of the embedding normalization on the 181M dense model (25B tokens): constraining every embedding vector to unit norm versus letting its norm vary. \figleft~LR sweep of the final loss for each mode. \figcenter~The loss over training for the various embedding modes. \figright~The relative embedding update over training.}
      \label{fig:embed-norm}
\end{figure}

\subsubsection{Magnitude Gains}
\label{sec:gains}

\textbf{Gain axis.} For the gains, we compare four different settings: a single scalar, a per-row vector, a per-column vector, and a combined per-row-and-column gain. All of them are initialized at 1, so they do not influence the initial model. With our results in \autoref{fig:gain-axis} (\panelleft), we see a noticeable improvement of adding learned magnitudes on top of spherical training; the combined row-and-column gain performs noticeably best. It therefore becomes our default setting. The results are a direct evidence that fine-grained magnitudes matter to the model: a single overall scale per matrix is not enough, and the gains let it amplify some rows and columns while damping others. We track the dynamics of learned gains in \autoref{sec:appendix-gain-dynamics} (\autoref{fig:gain-dynamics}) and find that they indeed spread out over a wide range across rows and columns over the course of training.

\textbf{Higher-rank gains.} Since the combined row-col gain is effectively an elementwise multiply by a rank-1 matrix (the outer product $\grow\gcol^\top$), it is natural to ask whether going higher-rank does even better; we take first steps in this direction in \autoref{sec:appendix-higherrank} (\autoref{fig:higherrank-gains}) and find no initial evidence that using rank-$k$ matrices improves the results.

\textbf{Gain parameterization.} Our results in \autoref{fig:gain-axis} (\panelcenter and \panelright) show that the parameterization matters little. We compare updating $\gamma$ directly (with and without a $10^{-5}$ floor), the exponential map $\gamma=e^{\widehat{\gamma}}$, and the smooth softplus map
\begin{equation}
\gamma = \varphi(\widehat{\gamma}) = \log\!\big(1 + e^{\widehat{\gamma}}\big).
\end{equation}
All four train stably and reach essentially the same loss: even updating $\gamma$ directly --- with no clamping and nothing to prevent a sign flip --- trains fine, and the loss curves are nearly indistinguishable. The softplus map, which keeps $\gamma > 0$ by construction and keeps the gradient smooth even around zero, gives a small but consistent edge over the others at the optimal LR, so we adopt it as our default.\footnote{Our initial experiments did see instability for small gains close to zero, which is what led us to explore positive parameterizations like softplus in the first place. We later traced this to a bug (a mismatch between how the gains were divided out and re-applied, attempting to prevent division by zero) rather than anything fundamental about updating $\gamma$ directly. With the bug fixed, all parameterizations train stably; softplus still comes out slightly ahead, so we keep it as our default.} We do not tune the gains' LR separately: each gain follows the learning rate of the matrix group it belongs to, and we find the loss to be very insensitive to it over more than an order of magnitude (\autoref{fig:lr-panels-combined} in \autoref{sec:appendix-lr-sweeps}).

\begin{figure}[t]
\centering
\begin{subfigure}{0.325\textwidth}\includegraphics[width=\textwidth]{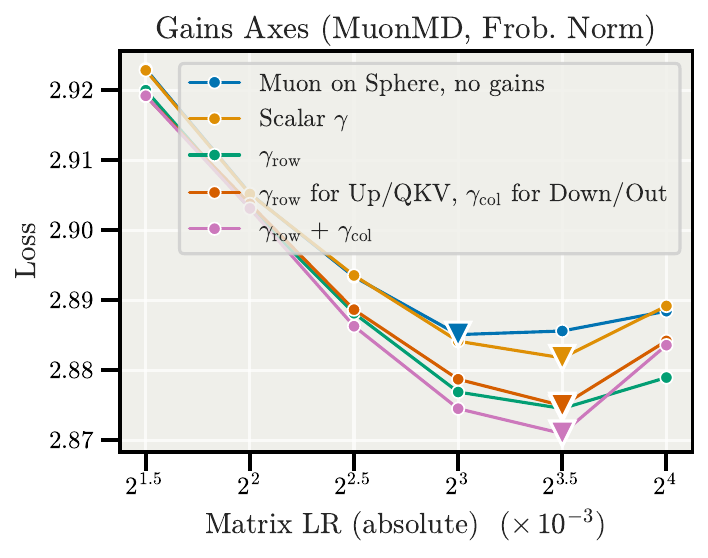}\end{subfigure}\hfill
\begin{subfigure}{0.325\textwidth}\includegraphics[width=\textwidth]{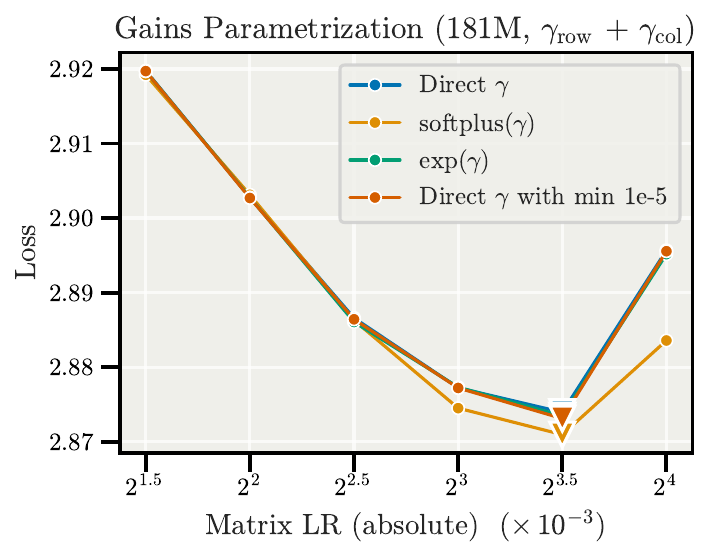}\end{subfigure}\hfill
\begin{subfigure}{0.325\textwidth}\includegraphics[width=\textwidth]{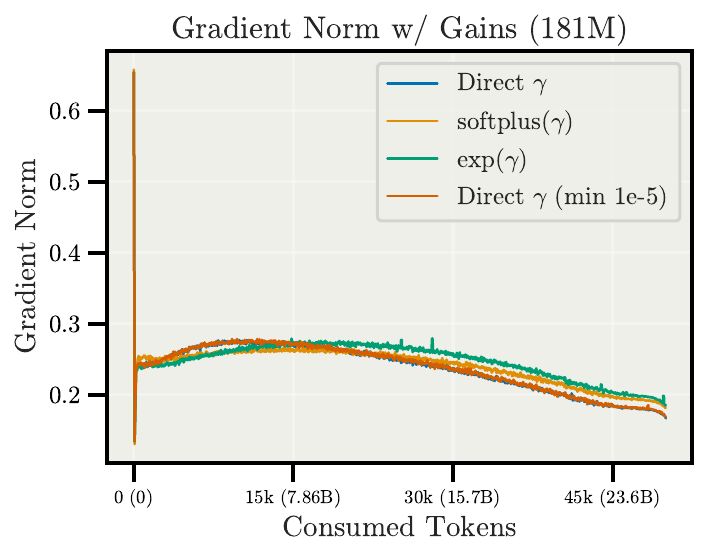}\end{subfigure}
\caption{\textbf{Adding learnable magnitude gains on top of spherical training helps noticeably; a combined per-row-and-column gain works best. The gain parameterization makes little difference, with softplus giving a minimal edge, and all parameterizations training stably.} On the 181M dense model (25B tokens). \figleft~LR sweep over gain modes (scalar, per-row, per-column, or both rows and columns). \figcenter~LR sweep over gain parameterizations: updating the gain directly (with and without a $10^{-5}$ floor), the exponential map, and the smooth softplus map. \figright~The gradient norm over training; all four parameterizations train stably with nearly indistinguishable gradient-norm curves.}
\label{fig:gain-axis}
\end{figure}

\subsubsection{Learning-Rate Transfer}
\label{sec:transfer}

Controlling the relative weight update directly through the spherical constraint comes with a crucial benefit: the optimal LR stays fixed as we scale the model in width, so it can be tuned once on a small model and reused on a much larger one. We are not the first to see this from a sphere / relative-update perspective: the same effect was shown with LionAR in earlier work~\citep{kosson2025weight}, and MuonH~\citep{muonh2026} and HyperP~\citep{ren2026rethinking} report it for the Frobenius-sphere constraint with Muon. We verify that it carries over to Muon with magnitude--direction decoupling, and propose a simple recipe for depth-transfer as well.

We sweep the matrix LR while scaling the model in width, depth, and both at once, and track the relative weight update and activation statistics underlying the transfer (\autoref{fig:transfer} and \autoref{fig:transfer-dynamics}).

\begin{figure}[t]
\centering
\begin{subfigure}{0.325\textwidth}\includegraphics[width=\textwidth]{sweep_exp2_width_transfer.pdf}\end{subfigure}\hfill
\begin{subfigure}{0.325\textwidth}\includegraphics[width=\textwidth]{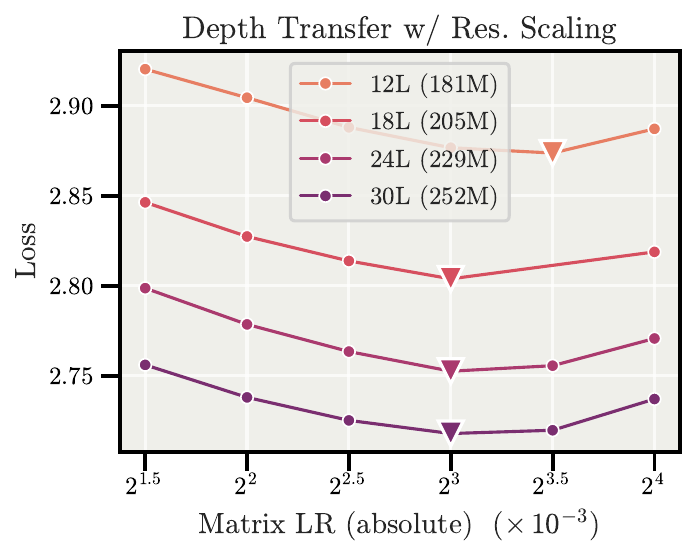}\end{subfigure}\hfill
\begin{subfigure}{0.325\textwidth}\includegraphics[width=\textwidth]{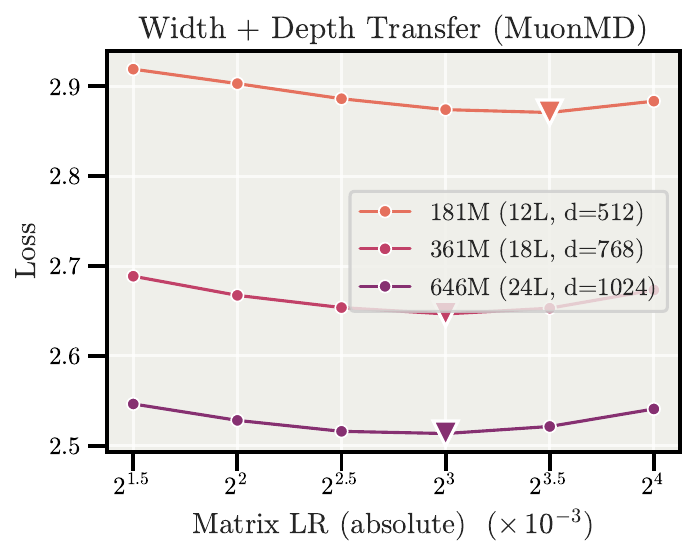}\end{subfigure}
\caption{\textbf{With Magnitude--Direction Decoupling the optimal matrix learning rate stays essentially fixed as the model grows, so it can be tuned once on a small model and reused.} Matrix-LR sweeps on dense models (from the 181M base) scaled across width \figleft~, depth \figcenter~, and width and depth \figright~ jointly. In each panel the optimal matrix LR stays roughly fixed across model sizes.}
\label{fig:transfer}

\medskip

\begin{subfigure}{0.325\textwidth}\includegraphics[width=\textwidth]{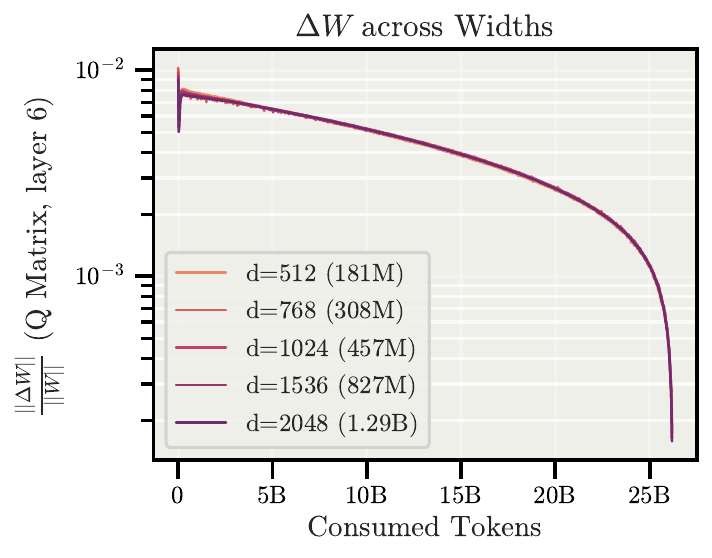}\end{subfigure}\hfill
\begin{subfigure}{0.325\textwidth}\includegraphics[width=\textwidth]{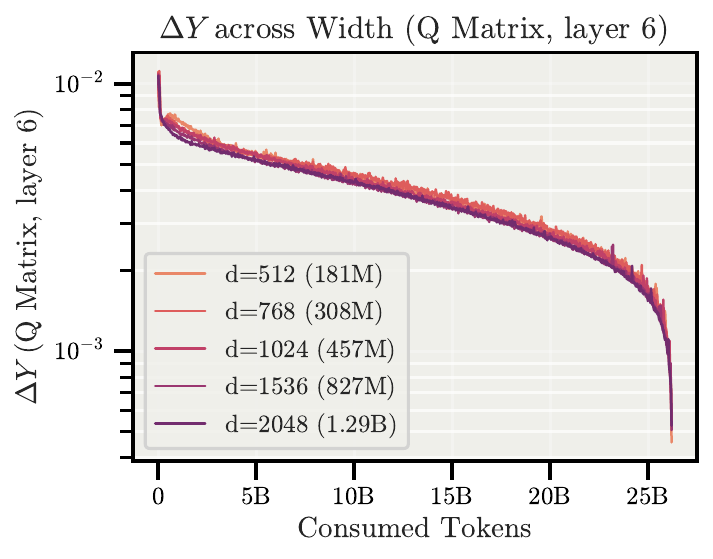}\end{subfigure}\hfill
\begin{subfigure}{0.325\textwidth}\includegraphics[width=\textwidth]{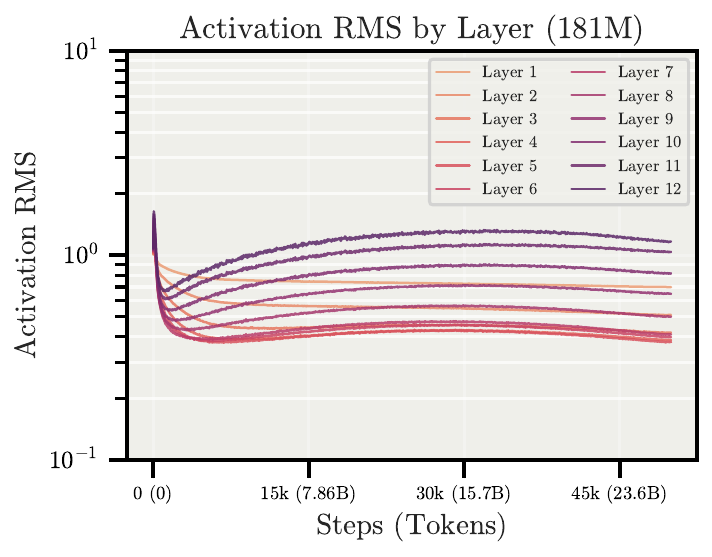}\end{subfigure}
\caption{\textbf{The LR transfer of \autoref{fig:transfer} arises because the relative weight update and the resulting activation changes are held constant as the model scales.} Training dynamics on the same dense models. \figleft~The relative weight update is precisely controlled across width. \figcenter~This in turn keeps the relative change in the layer's output stable. \figright~Across depth, the per-layer activation RMS stays well-behaved across the layers of a single model.}
\label{fig:transfer-dynamics}
\end{figure}

\textbf{Results.} The optimal matrix LR is essentially flat across \textbf{width} (\autoref{fig:transfer}, \panelleft), following from the precise control of relative weight updates on the sphere (\autoref{fig:transfer-dynamics}, \panelleft and \panelcenter). Transfer across \textbf{depth} relies on the fixed block-output scale $\alpha=\frac{1}{L}$ interplaying with the gains of the block-output RMSNorm. Other ways to downscale the residual work just as well (\autoref{sec:appendix-depth}, \autoref{fig:depth-scaling}); the main observation is that depth scaling needs no other tricks (\autoref{fig:transfer}, \panelcenter). Combined, the same transfer holds when scaling both width and depth jointly (\autoref{fig:transfer}, \panelright).

Note that we transfer only the matrix LR; the embedding and output-layer LRs are held fixed across model sizes rather than scaled as they require a different axis (since trained with Adam). Additionally, the transfer explored is in width and depth at \emph{fixed} batch size and training length. Transfer across batch size and token budget is a separate question, for which we show first experiments in the MoE Section (\autoref{sec:moe}).
For completeness, the AdamW and Muon baselines (changing only the matrix LR) are in \autoref{fig:transfer-baselines}.

\subsubsection{Learning-Rate Schedules on the Sphere}
\looseness-1
Since the relative update follows the schedule directly on the sphere (no weight decay, no equilibrium to drift to), the shape of the LR matters more than in standard training. We compare the established recipe of a Warmup-Stable-Decay (WSD) schedule ~\citep{hu2024minicpm} with a 20\% cooldown in a \texttt{1-sqrt} shape~\citep{hagele2024scaling} against a simple linear schedule, and look at the weight-change dynamics in \autoref{fig:lr-decay}.

What we see is that the established recipe no longer matches a full annealing on the sphere; in fact it is far from it. In standard, unconstrained training the weight norm grows over the stable phase, inducing an \emph{implicit} LR decay even at a constant nominal LR, so a WSD schedule decays more than its nominal shape suggests. On the sphere there is no such norm growth, and the relative update tracks the schedule exactly (\autoref{fig:lr-decay}, \panelright). The optimal on-sphere schedule remains an open question, but consistent, gradual annealing appears to be a key ingredient.

\subsubsection{Warmup-Free and Continual Training}

Decoupling removes the need for warmup: the large, destabilizing updates of the first few steps never appear, since the relative update is regulated from step one. In fact, dropping warmup even \emph{improves} the loss rather than merely matching it (\autoref{fig:warmup}, \panelleft): with warmup the early steps are wasted at a reduced LR even though training is already stable, so removing it puts them to productive use. This gain is even larger than for standard Muon.

The analogous question arises when training resumes from a checkpoint rather than from initialization (as in SFT or other post-training). To de-risk this, we run a re-warming experiment on a 150M model (\autoref{fig:warmup}, \panelcenter and \panelright): both the loss and the gradient norm behave well when we resume and re-warm, even when we reuse the optimizer state and use \emph{no warmup}. Staged and continual training therefore do not seem to be a problem. The impact of decoupling on properties such as \emph{sharpness} is an interesting question for further study \citep{springer2025overtrained,watts2026sharpness}.

\begin{figure}[t]
\centering
\begin{subfigure}{0.325\textwidth}\includegraphics[width=\textwidth]{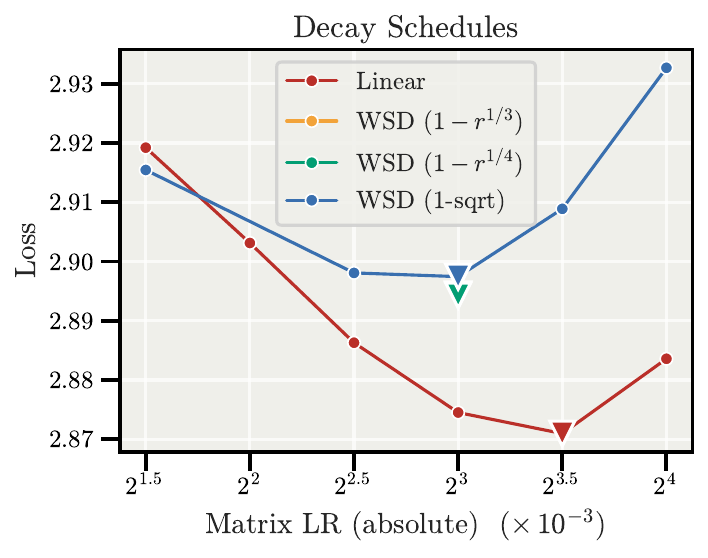}\end{subfigure}\hfill
\begin{subfigure}{0.325\textwidth}\includegraphics[width=\textwidth]{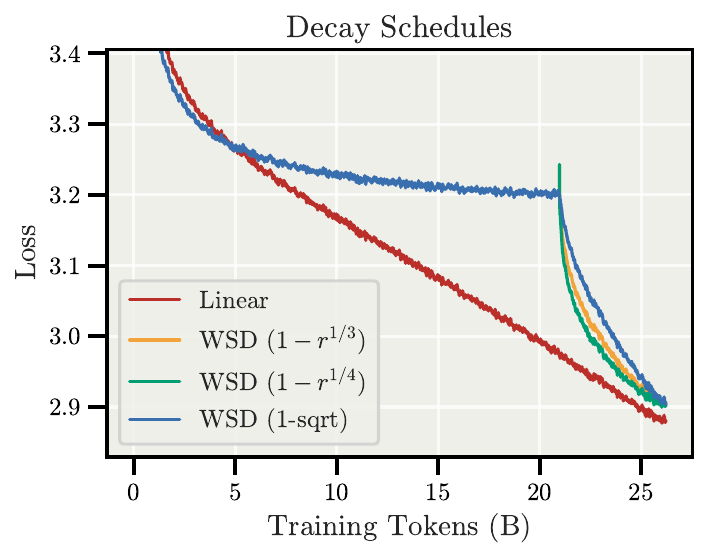}\end{subfigure}\hfill
\begin{subfigure}{0.325\textwidth}\includegraphics[width=\textwidth]{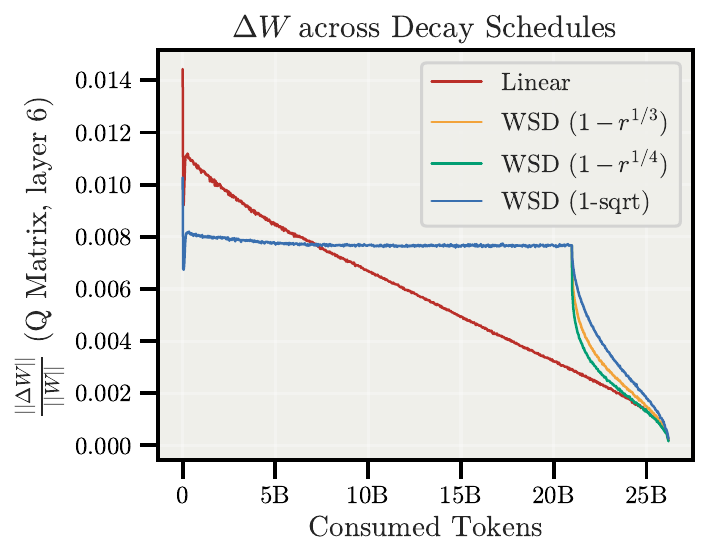}\end{subfigure}
\caption{\textbf{On the sphere, the relative weight update follows the learning-rate schedule directly, so the shape of the decay matters more than it does under weight decay.} Comparison of a Warmup-Stable-Decay (WSD) schedule against a simple linear decay on the 181M dense model. \figleft~LR sweep comparing WSD and linear decay. \figcenter~The corresponding loss curves. \figright~The relative weight update for the attention query projection, which follows the decay shape directly (the distortion at the very end is an artifact of the very low LRs during the final cooldown steps).}
\label{fig:lr-decay}

\medskip

\begin{subfigure}{0.325\textwidth}\includegraphics[width=\textwidth]{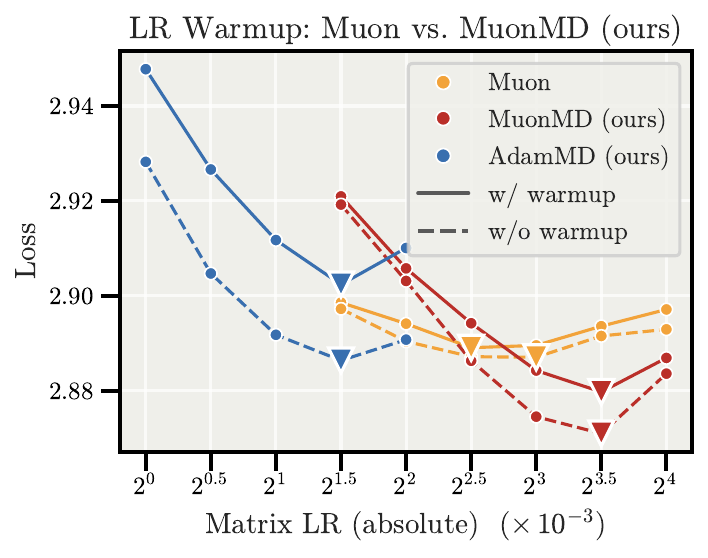}\end{subfigure}\hfill
\begin{subfigure}{0.325\textwidth}\includegraphics[width=\textwidth]{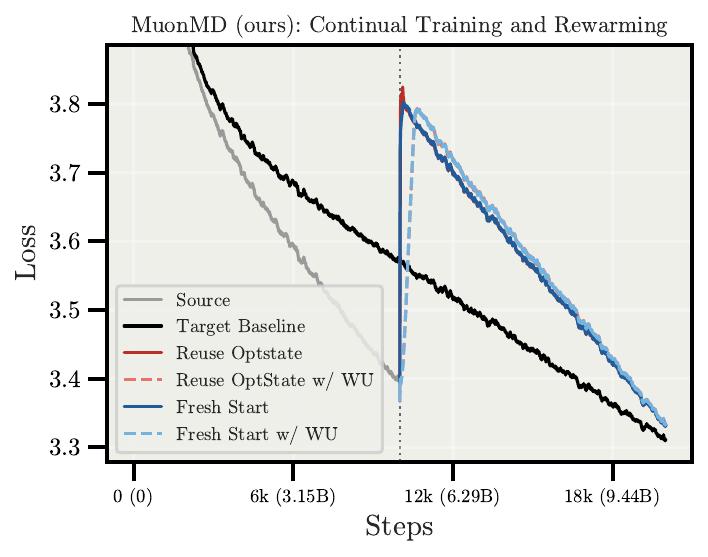}\end{subfigure}\hfill
\begin{subfigure}{0.325\textwidth}\includegraphics[width=\textwidth]{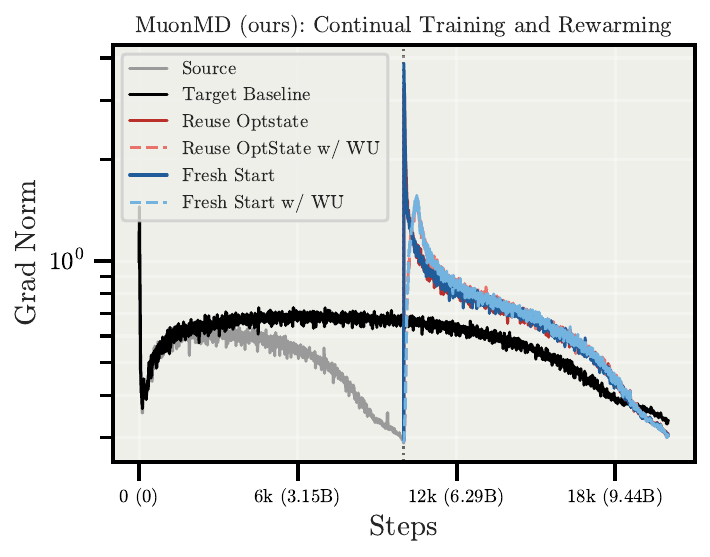}\end{subfigure}
\caption{\textbf{Decoupling and fixing weight norms removes the need for warmup, and dropping it improves the loss, both from the beginning and when resuming training from a checkpoint.} \figleft~LR sweep with and without warmup on the 181M model, showing dropping warmup improves the loss. \figcenter~Loss curves for re-warming runs on a 150M model. \figright~The gradient norm over the same re-warming runs, confirming training stays stable.}
\label{fig:warmup}
\end{figure}

\subsection{Scaling to Large Mixture-of-Experts Models}
\label{sec:moe}

\textbf{Setup.} We now turn to Mixture-of-Experts (MoE) transformers to check that our improvements persist at scale. We use a DeepSeekMoE-style architecture~\citep{dai2024deepseekmoe}: sizes range from 1.2B--6.7B total and 270M--810M active parameters, at a fixed 6\% non-embedding sparsity with 64 experts (1 shared) and top-2 routing. Unlike the dense models above, these do not use RMSNorms after the attention and MLP blocks. We train on the Apertus 1.0~\citep{swissai2025apertus} phase-5 data mixture (DCLM-edu, FineWeb-2 HQ), a good mixture for verifying routing in multilingual settings. The full MoE configurations and hyperparameters are in \autoref{sec:appendix-setup}.

The recipe has three steps: find the optimal LR on a small base model, transfer it to larger models with a scaling rule, and then scale up and compare.

\textbf{Step 1: find the optimal LR.} We fix a base model at 270M-active / 1.2B-total and train for ${\sim}15$B tokens (28k steps), sweeping the matrix LR for each optimizer individually (\autoref{fig:moe}, \panelleft). Here the remaining LRs (embeddings, LM head, gains) are all held fixed at $10^{-3}$ with Adam, and the standard methods use a weight decay of $0.1$, so each optimizer is tuned with the same budget on a shared base. For Muon we use a shape-scaling factor of $\max\!\big(1,\ \sqrt{\frac{\dout}{\din}}\big)$; the lower bound of 1 matters so the router is not given a much downscaled LR. For MuonMD we additionally normalize the routers along the expert axis (rows), otherwise following the earlier recipe of scaling the update to match the weight norm.\footnote{As for plain Muon, the router gets no upscaling under MuonMD, but a scale factor of $1$, leaving its LR unscaled.}

\textbf{Step 2: transfer it to larger models.} With the optimal matrix LR fixed from the base sweep, we scale up \emph{without re-tuning}. For the AdamW and Muon baselines we follow the Complete(d)P~\citep{dey2025completep,mlodozeniec2025completed} parametrization to set the LR and weight decay for all parameter groups across model width and training length. For MuonMD the recipe is simpler: because the sphere constraint already gives LR transfer across width (\autoref{sec:transfer}), we need no width multiplier at all and only have to account for the training length. There we scale LRs by $1/T^{0.25}$ ($T$ the token-count scaling factor), gentler than Complete(d)P's $1/T^{0.5}$ on the nominal LR, motivated by intuitions from rotational equilibrium~\citep{kosson2023rotational}, where the effective LR $\sqrt{\eta\lambda}$ governs the dynamics.
\footnote{The right exponent for scaling the LR with training length is still unsettled; other works report values around $0.3$ (e.g. HyperP~\citep{ren2026rethinking}). A principled choice for the sphere setting, where the LR directly sets the relative update, remains open. We briefly verified that $0.25$ works better than $0.5$ for MD training, but have not studied it exhaustively yet.}

\textbf{Step 3: scale up and compare.} Putting it together, we scale to the large MoEs and compare against the baselines transferred from the tuned base (\autoref{fig:moe}), varying dataset size $D$ from 7.5B--44B tokens. For the scaling law we plot loss against compute measured as \emph{non-embedding active-parameter} FLOPs, i.e. $6\,N_\text{act}\,D$ with $N_\text{act}$ the number of active non-embedding parameters. For the third experiment we reuse the 270M-active / 1.2B-total base config and increase the batch size by $k$ while scaling the LR by $\sqrt{k}$~\citep{malladi2022sdes}.

\begin{figure}[t]
\centering
\begin{subfigure}{0.325\textwidth}\includegraphics[width=\textwidth]{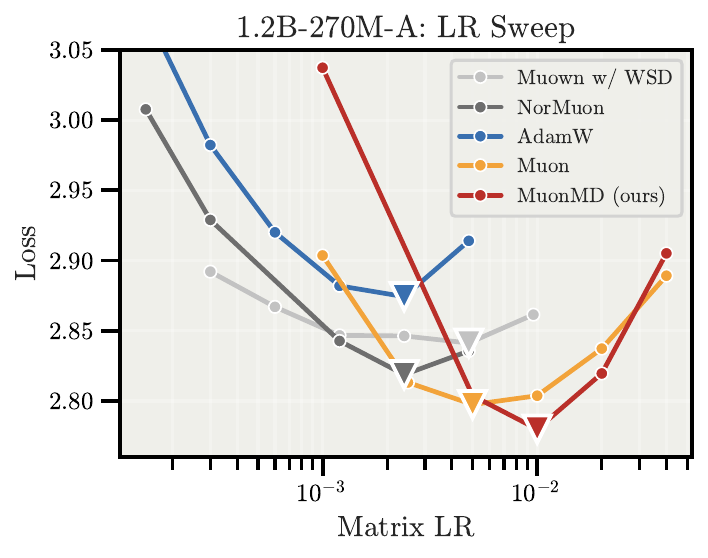}\end{subfigure}\hfill
\begin{subfigure}{0.325\textwidth}\includegraphics[width=\textwidth]{scaling-laws-scaling-frontier-power-shared-flops-ne.pdf}\end{subfigure}\hfill
\begin{subfigure}{0.325\textwidth}\includegraphics[width=\textwidth]{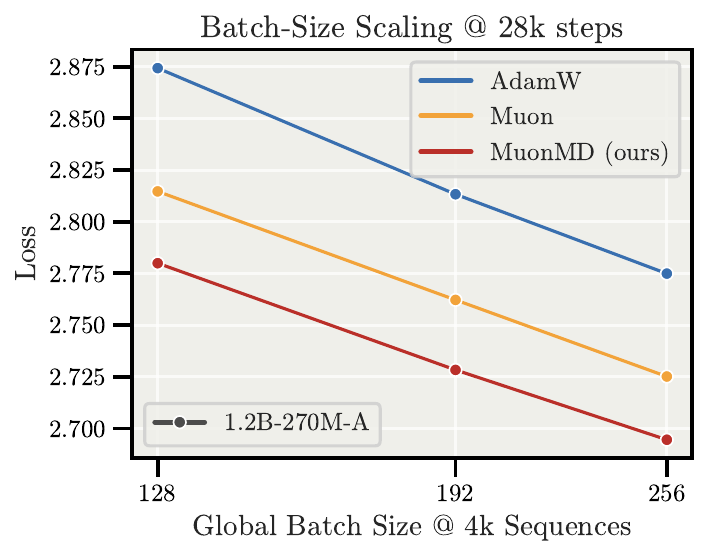}\end{subfigure}
\caption{\textbf{The gains from decoupling persist at scale: on large MoEs, MuonMD beats well-tuned Muon and AdamW and reaches AdamW's loss with roughly $2\times$ less compute.} DeepSeekMoE-style models (270M--810M active parameters). \figleft~LR sweep of the base optimizers on the 270M-active base. \figcenter~Scaling law of loss vs.\ compute (non-embedding active-parameter FLOPs), where the improvement holds across a wide range of compute. \figright~Batch-size scaling as a function of the global batch size. The improvement also carries over to downstream benchmarks (\autoref{sec:appendix-evals}).}
\label{fig:moe}
\end{figure}

\textbf{Results.} The gains from decoupling persist at scale and in the MoE setting, where MuonMD improves on tuned Muon and AdamW (\autoref{fig:moe}, \panelleft). When scaling models and training lengths, MuonMD stays ahead across the full range of compute we tested, reaching AdamW's loss with roughly $2\times$ less compute (\autoref{fig:moe}, \panelcenter). This advantage holds across global batch sizes (\autoref{fig:moe}, \panelright), and it carries over from the pretraining loss to downstream benchmarks: on the largest MoE (6.7B total / 810M active), MuonMD attains the best task average and leads on four of five zero-shot evaluations (\autoref{sec:appendix-evals}).

\textbf{Limitations.} Our protocol is principled, but certainly imperfect.
\begin{itemize}
\item The Adam-group LRs were not separately tuned, and the initial values might be suboptimal (which would spill into other configurations).
\item We did not verify that Complete(d)P holds exactly in our setting or how it changes with Muon, and we did not apply the depth scaling rules. Still, it represented the most rigorous approach accounting for the different axes of scaling without spending large compute on tuning.
\item We initially kept a short warmup for Muon, assuming it helps the Adam-optimized parameters not held on a sphere, but this no longer seems to be the case (\autoref{fig:warmup}).
\item When varying the training tokens across models, we did not change the batch size or iteration count, which might change the compute-optimal scaling.
\end{itemize}

\section{Related Work}
\label{sec:related}

The ideas behind magnitude--direction decoupling have surfaced, in different forms, across a rapidly growing body of recent and concurrent work, with many of these threads developed independently and in parallel. We try to connect them here, and have aimed to ablate the choices directly in our experiments (\autoref{sec:details}). We discuss the most directly related work below and give a fuller account in \autoref{sec:appendix-relwork}.

\textbf{Most recent and directly related.} AdamH/MuonH~\citep{muonh2026} constrain the hidden weights of LLMs to a fixed Frobenius sphere and drop weight decay, like our direction update. HyperP~\citep{ren2026rethinking} builds on top of this and investigates how to achieve LR transfer across width, depth, training tokens, and MoE granularity. However, both keep the norms fixed but add no learnable magnitude gains (relying on normalization layer gains). In constrast, learnable Multipliers~\citep{velikanov2026learnable} and Scale Vectors~\citep{wang2026negligible} both learn explicit scales, similar to us; the former adding scalar/per-row/per-column multipliers to each matrix, the latter studying the scale vectors in normalization layers and proposing a magnitude--direction reparameterization. Neither holds the weights at a fixed sphere norm.
Concurrently and very close to our work, Muown~\citep{lion2026muown} splits each weight into a per-row magnitude and a directional factor --- motivated by Muon's tendency to let the spectral norm drift upward --- updating the magnitude with Adam and the direction with Muon, a separation much like ours but per row. Originally, it let the norm grow at each step, which gives an implicit LR decay for growing weights; the direct follow-up~\citep{huebler2026angularmuown} independently arrives at the same idea as our work, i.e., projecting each weight onto a row sphere to explicitly control the angular update, and supports it with Riemannian gradient theory. Unlike both, which are tied to Muon, our motivation is optimizer-agnostic, and we additionally ablate relevant empirical choices, such as the normalization axis and more expressive gains.

\textbf{Broader context.} A long line of work constrains weights (and sometimes updates) to a sphere or matrix manifold during pretraining --- on the per-vector or Frobenius sphere~\citep{loshchilov2024ngpt,franke2025learning,fu2025nemotron,gu2026mano} or on the spectral norm~\citep{xie2026controlled,bernstein2025modular,newhouse2025training,xu2026width} --- differing mainly in \emph{which} norm is fixed and whether a magnitude is added back. In our final recipe, we fix the softer Frobenius norm and add learnable gains inside the optimizer. A related thread controls the update size \emph{relative} to the weight without an explicit split, building on the ``effective learning rate'' of scale-invariant weights under normalization~\citep{vanlaarhoven2017l2,wan2021spherical,kodryan2022training,kosson2023rotational,liu2021learning,you2017large,karras2023analyzing}, and a parallel one transfers hyperparameters across scale via $\mu$P and its successors~\citep{yang2022tensor,shigida2026learning,mlodozeniec2025completed,dey2025completep}. 
The most direct classic ancestor of our gains is Weight Normalization~\citep{salimans2016weight}, which reparameterizes $w=(g/\lVert v\rVert)\,v$ as a learnable magnitude times a direction, though without a fixed-norm constraint or a separate LR for the direction; this and other reparameterization and normalization schemes~\citep{liu2018decoupled,qiao2019micro,miyato2018spectral} target conditioning or stability rather than the magnitude--direction interference we address.

\section{Discussion}
\label{sec:discussion}
Magnitude--direction decoupling is a simple change that pays off across optimizers and scales. While the additional logic required in the optimizer step adds a few more operations, we find that the overhead introduced relative to the total training time remains modest as the model size scales up in practical distributed settings (\autoref{sec:appendix-efficiency}). Connecting to existing work on spherical training, our ablations clarify what matters: fixing the direction to a sphere buys most of the gain (while the exact normalization axis is secondary), and adding \emph{learnable} per-row and per-column gains brings crucial improvements. 

More broadly, MD training invites us to rethink how we parametrize and constrain networks from their training dynamics. It also opens questions on the optimal on-sphere schedule, what the learned magnitudes do (\autoref{sec:appendix-gain-dynamics}), and understanding the impact on loss-landscape sharpness, downstream behaviour (in particular RL), or low-precision training. 

\clearpage
\ifanon\else
\subsubsection*{Acknowledgments}
This work used compute from the Swiss AI Initiative on the Alps cluster under the Apertus initiative. We thank Fabian Schaipp, Mikhail Gorbunov, and Skander Moalla for helpful discussions, and Andres Nowak for help with the evaluation suite.
\fi

\bibliography{main}
\bibliographystyle{tmlr}
\newpage

\appendix

\section{Full Optimizer Step with Row-and-Column Gains}
\label{sec:full-algorithm}

\autoref{alg:md-step} in the main text gives the optimizer step for a single scalar gain. \autoref{alg:md-step-full} spells out the general case we use as our default: separate per-row and per-column gains $\grow,\gcol$ with the softplus reparameterization of \autoref{sec:optimizer}. The scalar version is recovered by tying both gains to a single shared scalar, and the per-row-only or per-column-only variants by fixing the other gain to $1$.

\begin{algorithm}[h]
\caption{Magnitude--Direction decoupled optimizer step with per-row and per-column gains. The fused weight is $W=\diag(\grow)\,\What\,\diag(\gcol)$ with positive gains $\grow=\varphi(\widehat{\grow})$, $\gcol=\varphi(\widehat{\gcol})$ obtained from raw gains through a smooth map $\varphi$ (e.g.\ softplus). Here $\mathrm{rowsum}$/$\mathrm{colsum}$ reduce a matrix over its columns/rows, $\odot$ is the elementwise product, and $\varphi'$ is applied elementwise.}
\label{alg:md-step-full}
\begin{algorithmic}[1]
\Require fused weight $W$, raw gains $\widehat{\grow}\in\R^{\dout},\ \widehat{\gcol}\in\R^{\din}$, gradient $G=\partial L/\partial W$, direction LR $\eta_W$, gain LR $\eta_\gamma$, gain map $\varphi$, fixed sphere radius $c_F$
\State $\grow \leftarrow \varphi(\widehat{\grow})$,\quad $\gcol \leftarrow \varphi(\widehat{\gcol})$ \Comment{materialize the positive gains}
\State $\What \leftarrow \diag(\grow)^{-1}\,W\,\diag(\gcol)^{-1}$ \Comment{recover the on-sphere direction}
\State $g_{\grow} \leftarrow \mathrm{rowsum}\big((\What \odot G)\,\diag(\gcol)\big)$ \Comment{row-gain gradient: sum over columns}
\State $g_{\gcol} \leftarrow \mathrm{colsum}\big(\diag(\grow)\,(\What \odot G)\big)$ \Comment{column-gain gradient: sum over rows}
\State $g_{\widehat{\grow}} \leftarrow g_{\grow} \odot \varphi'(\widehat{\grow})$,\quad $g_{\widehat{\gcol}} \leftarrow g_{\gcol} \odot \varphi'(\widehat{\gcol})$ \Comment{backprop through the gain map}
\State $G_{\What} \leftarrow \diag(\grow)\,G\,\diag(\gcol)$ \Comment{direction gradient $\partial L/\partial \What$}
\State $\What \leftarrow \OptStep\big(\What,\, G_{\What},\, \eta_W\big)$ \Comment{any (normalized) matrix optimizer (Adam / Muon / \dots)}
\State $\What \leftarrow c_F \, \What \,/\, \lVert \What\rVert$ \Comment{project back onto the sphere of fixed radius $c_F$}
\State $\widehat{\grow} \leftarrow \AdamStep\big(\widehat{\grow},\, g_{\widehat{\grow}},\, \eta_\gamma\big)$,\quad $\widehat{\gcol} \leftarrow \AdamStep\big(\widehat{\gcol},\, g_{\widehat{\gcol}},\, \eta_\gamma\big)$ \Comment{step the raw gains (own LR)}
\State $W \leftarrow \diag(\varphi(\widehat{\grow}))\,\What\,\diag(\varphi(\widehat{\gcol}))$ \Comment{reassemble for the next forward}
\end{algorithmic}
\end{algorithm}

\section{Experimental Setup Details}
\label{sec:appendix-setup}

This section gives the full architecture and training details for both the dense ablations (\autoref{sec:dense}) and the large MoE experiments (\autoref{sec:moe}). All code is a fork of Megatron-LM~\citep{shoeybi2019megatron}; the magnitude--direction decoupling, the per-axis gains, and the Muon/orthogonalized updates are implemented inside the optimizer so the model always sees a single fused weight tensor (\autoref{alg:md-step-full}).\anon{}{\footnote{Our code is available for the dense models at \url{https://github.com/haeggee/Megatron-LM/tree/gainz} and for the MoE experiments at \url{https://github.com/haeggee/Megatron-LM/tree/feat/scaling-sweeps}. Note that in this research codebase, the optimizer is called \texttt{master}, which was the development name for our ablations.}} All runs are in \texttt{bf16}. Training was carried out \anon{}{on the Alps cluster at the Swiss National Supercomputing Centre (CSCS), }on GH200 nodes (4 GPUs/node), in pure data-parallel with the distributed optimizer. The largest MoEs optionally shard experts with expert-parallelism set to 4.

\subsection{Dense Models}
\label{sec:appendix-setup-dense}

\textbf{Architecture.} The dense models are GPT-style transformers with RoPE~\citep{su2021roformer} (base $\theta=5\!\times\!10^5$), SwiGLU MLPs~\citep{shazeer2020glu}, RMSNorm~\citep{zhang2019rmsnorm} ($\epsilon=10^{-5}$), GQA~\citep{ainslie2023gqa} with the number of key/value groups set to half the number of attention heads, a fixed head-dimension of $128$, and QK-RMSNorm~\citep{dehghani2023scaling,wortsman2023small}. We use Sandwich Norm~\citep{ding2021cogview,kim2025peri} (an extra RMSNorm on each block output) together with a fixed block-output scale $\alpha=\frac{1}{L}$. Embeddings and the output head are \emph{untied}. Matrix parameters are initialized with standard deviation $\frac{1}{\sqrt{d}}$, and the embeddings are upscaled by $\sqrt{d}$ so the residual stream has unit RMS at the input. The full grid is given in \autoref{tab:dense-arch}: a base model ($d=512,L=12$, 181M parameters), a width sweep at fixed depth, a depth sweep at fixed width, and a joint width-and-depth sweep. The tokenizer is the Apertus v1 tokenizer ($\sim$131k vocabulary), which together with the untied input/output embeddings dominates the parameter count at the smaller sizes.

\begin{table}[t]
\centering
\small
\caption{\textbf{Dense model configurations.} All models use head-dimension $128$, GQA with $\text{KV groups}=\text{heads}/2$, SwiGLU, RMSNorm, QK-norm, Sandwich Norm, untied embeddings, and sequence length $4096$. The 181M base ($d=512,L=12$) is shared across all three sweeps. ``Params'' is the total count including the untied input/output embeddings.}
\label{tab:dense-arch}
\begin{tabular}{llccccc}
\toprule
Sweep & Params & $d$ & $L$ & Heads & KV groups & $d_{\text{ff}}$ \\
\midrule
Base                & 181M  & 512  & 12 & 4  & 2 & 2048 \\
\midrule
\multirow{4}{*}{Width ($L{=}12$)}
                    & 307M  & 768  & 12 & 6  & 3 & 3072 \\
                    & 457M  & 1024 & 12 & 8  & 4 & 4096 \\
                    & 827M  & 1536 & 12 & 12 & 6 & 6144 \\
                    & 1.29B & 2048 & 12 & 16 & 8 & 8192 \\
\midrule
\multirow{3}{*}{Depth ($d{=}512$)}
                    & 205M  & 512  & 18 & 4  & 2 & 2048 \\
                    & 229M  & 512  & 24 & 4  & 2 & 2048 \\
                    & 252M  & 512  & 30 & 4  & 2 & 2048 \\
\midrule
\multirow{2}{*}{Width$+$Depth}
                    & 361M  & 768  & 18 & 6  & 3 & 3072 \\
                    & 646M  & 1024 & 24 & 8  & 4 & 4096 \\
\bottomrule
\end{tabular}
\end{table}

\textbf{Data and batching.} Dense models are trained on a FineWeb-Edu~\citep{penedo2024fineweb} subset at sequence length $4096$, with a global batch size of $128$ sequences ($\sim$$0.52$M tokens). The base model is trained for $50$k steps ($\sim$$25$B tokens) --- deliberate strong overtraining (Chinchilla sense~\citep{hoffmann2022training}) to expose longer-horizon dynamics. The continual / re-warming experiment (\autoref{fig:warmup}) uses a 150M model ($d=512$, $L=8$, head-dimension $64$).

\textbf{Optimizer and learning rates.} The matrix-parameter optimizer is AdamW or Muon, optionally under MD decoupling. The Adam-managed groups (embeddings, LM head, norm gains, magnitude gains) use $(\beta_1,\beta_2)=(0.9,0.99)$ and $\epsilon=10^{-8}$; Muon uses a momentum of $0.95$ with Nesterov and $5$ Newton--Schulz iterations. We \emph{fix} the embedding LR at $3\!\times\!10^{-3}$ and the output-layer LR at $10^{-3}$, let the magnitude gains follow the matrix LR, and sweep only the \emph{matrix LR} per method, so every method is tuned with the same budget. The standard AdamW/Muon baselines use decoupled weight decay $0.1$; the MD variants use none (the weights are already norm-constrained). Gradients are clipped to global norm $1.0$. The schedule is a linear decay to $\text{min-LR}=10^{-8}$. The AdamW baseline uses $1000$ steps of warmup throughout; Muon and the MD variants are run warmup-free in the headline comparison (the warmup ablation is in \autoref{fig:warmup}). The schedule ablation (\autoref{fig:lr-decay}) compares this against a WSD schedule~\citep{hu2024minicpm,hagele2024scaling,schaipp2025surprising,dremov2025training} with a $20\%$ cooldown in the negative-square-root (``1-sqrt'') shape. For Muon's scale factor we sweep the conventions in \autoref{fig:muon-scale-modes}: the plain-Muon headline runs use $\sqrt{\frac{\dout}{\din}}$ (tied for best with shape scaling $\max(1,\sqrt{\frac{\dout}{\din}})$, and noticeably better than the RMS-matching factor), while MuonMD uses the factor $\sqrt{\max\!\big(\frac{\dout}{\din},\, \frac{\din}{\dout}\big)}$; the latter we here coin ``shape up'' for simplicity. Its inspiration is matching the RMS of the weight norm under our initialization.

\subsection{Mixture-of-Experts Models}
\label{sec:appendix-setup-moe}

\textbf{Architecture.} The MoE models follow a DeepSeekMoE-style design~\citep{dai2024deepseekmoe}: $64$ routed experts with top-$2$ routing plus $1$ always-on shared expert at half a routed expert's width, with the first layer(s) dense and the remainder MoE ($\sim$$5\%$ dense layers). They share the dense models' backbone conventions (RoPE $\theta=5\!\times\!10^5$, SwiGLU, RMSNorm, GQA, QK-norm, untied embeddings, $\frac{1}{\sqrt{d}}$ init, $\sqrt{d}$ embedding upscaling) but use an attention head-dimension of $64$ and --- unlike the dense models --- \emph{no} post-attention/post-MLP (Sandwich) norm, because this experiment was run in a separate Megatron fork. The four rungs (\autoref{tab:moe-arch}) span 1.2B--6.7B total and 270M--1.5B active parameters at a fixed $\sim$$6\%$ non-embedding sparsity, an iso-sparsity proxy for much larger production MoEs. The tokenizer is the same Apertus v1 tokenizer as the dense models (vocabulary padded to a multiple of $128$).

\begin{table}[t]
\centering
\small
\caption{\textbf{MoE model configurations.} DeepSeekMoE-style: $64$ routed experts, top-$2$ routing, $+1$ shared expert at half a routed expert's width, $\sim$$6\%$ non-embedding sparsity. Head-dimension $64$, sequence length $4096$, global batch $128$. $L$ is given as (dense $+$ MoE) layers; $d_{\text{ff}}^{\text{moe}}$ is the per-routed-expert FFN width (the shared expert is half of it).}
\label{tab:moe-arch}
\begin{tabular}{lccccc}
\toprule
Active / Total & $L$ (dense$+$MoE) & $d$ & Heads & KV groups & $d_{\text{ff}}^{\text{moe}}$ (shared) \\
\midrule
0.27B / 1.2B  & $1+13$ & 768  & 12 & 4 & 512 (256) \\
0.41B / 2.5B  & $1+19$ & 1024 & 16 & 4 & 576 (288) \\
0.58B / 4.1B  & $1+21$ & 1280 & 20 & 4 & 704 (352) \\
0.81B / 6.7B  & $1+23$ & 1536 & 24 & 8 & 896 (448) \\
\bottomrule
\end{tabular}
\end{table}

\textbf{Routing.} We use DeepSeek-V3-style routing~\citep{deepseekai2024v3}: sigmoid gating, an auxiliary-loss-free per-expert bias~\citep{wang2024auxlossfree} (selection bias updated at rate $10^{-3}$, not the gate weights), a small complementary sequence-wise auxiliary load-balancing loss (coefficient $10^{-3}$), top-$k$ renormalization scaled by $2.5$, and the router logits computed in \texttt{fp32}. This routing policy is held invariant across the whole ladder; only the expert \emph{geometry} (\autoref{tab:moe-arch}) changes per size.

\textbf{Data and batching.} MoE models are trained on the Apertus 1.0~\citep{swissai2025apertus} phase-5 data mixture (DCLM-edu and FineWeb-2 high-quality multilingual~\citep{messmer-etal-2025-enhancing}), a good mixture for verifying routing in the multilingual setting, at sequence length $4096$ with a global batch of $128$ sequences. The base sweep trains at 270M-active / 1.2B-total for $\sim$$15$B tokens ($3{,}584{,}000$ samples, $\sim$$28$k steps); the scaling-law runs sweep the token budget over $\{7.5,\,15,\,23,\,44\}$B, with budgets placed along a $\sim$$55$ tokens/active-parameter diagonal.

\textbf{Optimizer, learning rates, and transfer.} As in the dense setup we fix the embedding, output, and gain LRs and tune the matrix LR. The base sweep at 270M active parameters (15B tokens) fixes the embedding / output / gains LR at $10^{-3}$, and the optima are matrix LR $2.4\!\times\!10^{-3}$ (AdamW), $5\!\times\!10^{-3}$ (Muon), and $10^{-2}$ (MuonMD). AdamW and Muon use $(\beta_1,\beta_2)=(0.9,0.95)$, decoupled weight decay $0.1$, and a warmup of $1000$ steps; MuonMD uses weight decay $0$ and no warmup. Muon uses momentum $0.95$ with Nesterov and the shape-scaling factor $\max(1,\sqrt{\frac{\dout}{\din}})$ (the lower bound of $1$ keeps the router from being given a downscaled LR); MuonMD instead uses the shape-up factor $\sqrt{\max\!\big(\frac{\dout}{\din},\, \frac{\din}{\dout}\big)}$ to match the weight-norm RMS, additionally normalizes the router rows along the expert axis, and uses the softplus gain parameterization with row-and-column gains and per-row embedding normalization. For MuonMD the attention and MLP output (residual-write) projections additionally use the scaled initialization $\sigma/\sqrt{2L}$ (with $\sigma=\frac{1}{\sqrt{d}}$, the standard GPT-2 residual scaling); since the sphere fixes each matrix at its initialization norm, this lowers the sphere-norm target of those projections (by $\frac{1}{\sqrt{2L}}$) and adjusts their shape-up factor accordingly. All runs use init std $\frac{1}{\sqrt{d}}$ and embedding multiplier $\sqrt{d}$, with a linear decay to $\text{min-LR}=10^{-5}$ (absolute floor per group), which we found to work well and better than a WSD schedule. The one exception is Muown, which was better with WSD and is therefore shown with a $20\%$ linear cooldown; \autoref{fig:moe-wsd} reports the same base sweep with the WSD-schedule runs added. The AdamW baseline clips gradients to global norm $1.0$; the Muon and MuonMD runs use no gradient clipping.

To transfer the tuned base LRs to larger models without re-tuning, we follow Complete(d)P~\citep{dey2025completep,mlodozeniec2025completed} for the AdamW and Muon baselines, scaling the matrix/embedding/output LRs by $1/k$ in width ($k=d/768$) and all LRs by $1/\sqrt{l}$ in training length ($l=$ tokens$/15$B), with weight decay scaled $\propto k/\sqrt{l}$. MuonMD needs no width multiplier (the sphere constraint already transfers across width, \autoref{sec:transfer}); its LRs are scaled only with length, by $1/l^{0.25}$ (gentler than Complete(d)P's $1/\sqrt{l}$ on the nominal LR), and its weight decay stays $0$. The batch-size experiment reuses the 270M-active base config and increases the global batch by a factor $k$ while scaling the LR by $\sqrt{k}$~\citep{malladi2022sdes}.

\textbf{Scaling-law fitting.} For the scaling-law plot (\autoref{fig:moe}, \panelcenter, and \autoref{fig:headline}, \panelcenter), each run is summarized by its \emph{tail loss}, the mean LM loss over its last $25$ training iterations, and placed at a compute of $C=6\,N_\text{act}\,D$ with $N_\text{act}$ the active \emph{non-embedding} parameter count and $D$ the number of training tokens. For each optimizer we take the compute-optimal lower envelope --- the runs that set a new record-low loss as compute grows --- and fit a power law $L(C)=A\,C^{-\alpha}$ to those frontier points. We assume an irreducible floor $E=0$, which is reasonable over our experimental compute range (where a nonzero floor is not separately identifiable); the fit is then a simple log--log linear regression. The heavily-undertrained 810M-active@7.5B point is plotted but excluded from the envelope and fit. The fitted exponents are nearly identical across the three optimizers ($\alpha\approx0.05$), so the improvement is essentially a downward level shift --- a smaller coefficient $A$ --- rather than a steeper slope: at any compute in range, Muon and especially MuonMD reach a lower loss, equivalently the same loss at less compute. We quantify this as the \emph{compute savings} relative to the AdamW baseline at a fixed target loss $L=2.635$ (within the fitted range), i.e.\ the ratio $C_\text{AdamW}(L)/C_\text{opt}(L)$ of compute needed to reach $L$, with confidence intervals from a nonparametric bootstrap ($500$ resamples of the frontier points, refitting each time): Muon reaches $L$ at $1.55\times$ less compute (CI $[1.49,1.63]$) and MuonMD at $2.01\times$ less compute (CI $[1.94,2.11]$).

\begin{figure}[t]
\centering
\includegraphics[width=0.45\textwidth]{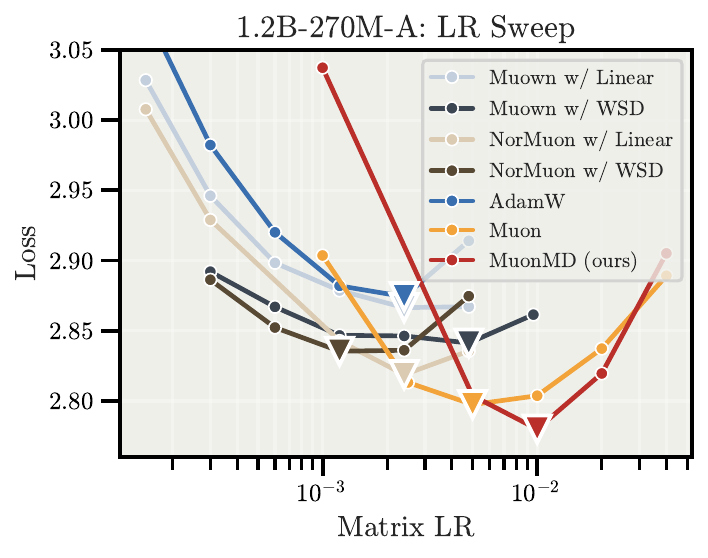}
\caption{\textbf{Additional LR sweeps comparing linear and WSD.} The 270M-active base LR sweep of \autoref{fig:moe} (\panelleft), shown here with the additional WSD-schedule runs for completeness. All optimizers use the linear decay as described in the main text, which we found to work best and better than WSD; the exception is Muown, which performs better with a WSD schedule (a $20\%$ linear cooldown) and is therefore plotted with that schedule in \autoref{fig:moe}.}
\label{fig:moe-wsd}
\end{figure}

\section{Downstream Evaluations}
\label{sec:appendix-evals}

Beyond the pretraining loss, we evaluate the largest MoE (6.7B total / 810M active, trained for 44B tokens) on a standard suite of zero-shot benchmarks: MMLU~\citep{hendrycks2021measuring}, HellaSwag~\citep{zellers2019hellaswag}, ARC-Challenge and ARC-Easy~\citep{clark2018think}, and WinoGrande~\citep{sakaguchi2021winogrande}, all via the LM Evaluation Harness~\citep{eval-harness}. We report the harness headline metric per task (length-normalized accuracy for HellaSwag and ARC, plain accuracy for MMLU and WinoGrande). \autoref{tab:moe-evals} gives the final-checkpoint numbers and \autoref{fig:moe-evals} the trajectory over the last third of training. The ranking mirrors the loss results: MuonMD attains the best task average and leads on four of the five benchmarks, with Muon second and AdamW third. MMLU sits near the chance level ($25\%$) for all three at this scale, as expected, and is essentially uninformative here.

\begin{table}[t]
\centering
\caption{Downstream evaluation of the 6.7B-total / 810M-active MoE at the end of training (44B tokens). Accuracy (\%); length-normalized for HellaSwag/ARC, plain for MMLU/WinoGrande. Best per column in bold.}
\label{tab:moe-evals}
\begin{tabular}{lcccccc}
\toprule
Optimizer & MMLU & HellaSwag & ARC-c & ARC-e & WinoGrande & Avg. \\
\midrule
AdamW  & \textbf{25.6} & 53.0 & 32.3 & 61.5 & 53.8 & 45.2 \\
Muon   & \textbf{25.6} & 55.9 & 35.0 & 63.7 & 58.0 & 47.6 \\
MuonMD & 23.3 & \textbf{57.0} & \textbf{36.6} & \textbf{67.1} & \textbf{59.6} & \textbf{48.7} \\
\bottomrule
\end{tabular}
\end{table}

\begin{figure}[t]
\centering
\includegraphics[width=\textwidth]{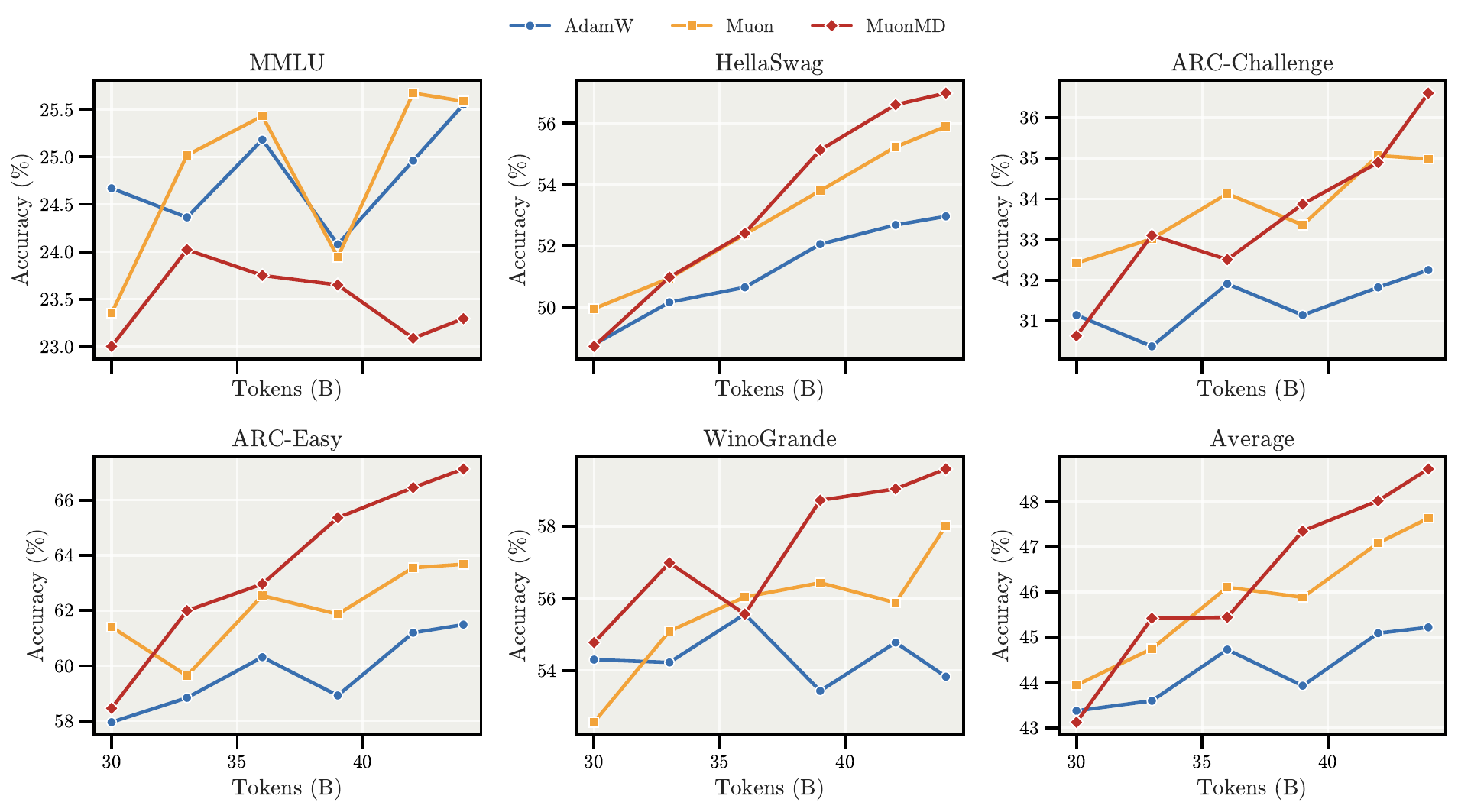}
\caption{\textbf{Downstream accuracy over training tokens for the 6.7B-total / 810M-active MoE.} Each panel tracks one benchmark's zero-shot accuracy against training tokens, with the 5-task average at the bottom right. MuonMD (red) leads the average throughout and ends ahead on all tasks except the near-chance MMLU. Metrics as in \autoref{tab:moe-evals}.}
\label{fig:moe-evals}
\end{figure}

\section{Additional Learning-Rate Ablations}
\label{sec:appendix-lr-sweeps}

This section collects the supporting learning-rate sweeps behind the choices we make throughout the paper: the fixed learning rates for the parameter groups that Adam manages, the sensitivity of the gains LR (\autoref{sec:details}), and the Muon scale factor.

\textbf{The different parameter-group learning rates.} Across all experiments we share a single recipe for the Adam-managed parameter groups --- the embeddings, the output (LM-head) layer, and the magnitude gains --- and sweep only the matrix LR per method, so that every method is tuned with the same budget. Since these groups are reused across setups without re-tuning, our goal here is twofold: (1) to verify that the fixed base learning rates sit in a good range, and (2) to probe the sensitivity of the gains LR specifically, as it governs the only group unique to our method. \autoref{fig:lr-panels-combined} establishes both on the 181M base model (25B tokens) across four panels. \figleft~The base Adam LR --- the shared learning rate of all Adam-managed groups (embeddings, output layer, and normalization layer gains) scaled together: holding the matrix LR at the AdamW optimum and sweeping this base LR gives an essentially flat curve, so our $10^{-3}$ choice is comfortably in range. \emph{(Center-left)}~The embedding LR: sweeping the matrix LR jointly with the embedding LR (ELR), the optimal matrix LR does not move and almost all ELR settings reach essentially the same loss, confirming $3\cdot10^{-3}$ as a safe default. \emph{(Center-right)}~The gains LR: under the softplus parameterization the loss is flat over more than an order of magnitude, so the magnitudes are remarkably insensitive to their LR. \figright~In the dense experiments the gains simply follow the matrix LR, but in the MoE experiments we instead fix the gains LR at $10^{-3}$; this panel verifies that choice, sweeping the matrix LR with the gains LR fixed and finding the optimal matrix LR again unchanged. Taken together, the loss is broad in every group except the matrix LR: as long as the shared Adam-group LRs are in a reasonable range, the matrix LR is the one hyperparameter worth sweeping per method.

\begin{figure}[t]
\centering
\includegraphics[width=\textwidth]{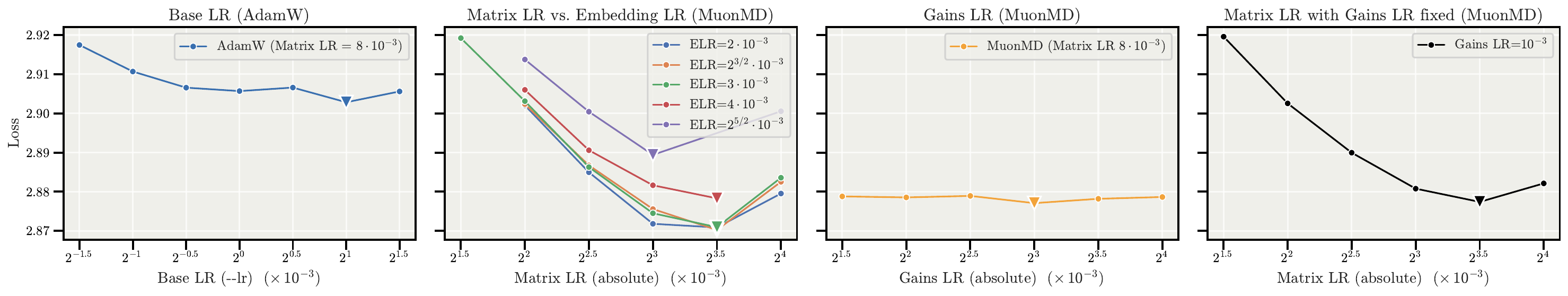}
\caption{\textbf{The matrix learning rate is the most important hyperparameter worth sweeping per method: the loss is broad in the other groups and essentially flat in the gains LR over more than an order of magnitude.} Learning-rate sweeps of the parameter groups held fixed in the main text (181M model, 25B tokens). \figleft~AdamW base LR --- the shared LR of all Adam-managed groups (embeddings, output layer, gains) --- with the matrix LR held fixed at its optimum; the curve is flat, so $10^{-3}$ is comfortably in range. \emph{(Center-left)}~MuonMD matrix LR swept jointly with the embedding LR (ELR): the optimal matrix LR is unchanged and all ELR settings coincide. \emph{(Center-right)}~MuonMD gains LR, which barely affects the loss over more than an order of magnitude under the softplus parameterization. \figright~MuonMD matrix LR at a gains LR fixed to $10^{-3}$ (as in the MoE experiments), where the optimal matrix LR is again unchanged.}
\label{fig:lr-panels-combined}
\end{figure}

\textbf{Muon scale factor.} Muon's orthogonalized update is typically multiplied by a shape-dependent factor, either to adapt the effective learning rate to the matrix shape or to match a desired update RMS, and several conventions exist: a unit-RMS-norm factor $\sqrt{\frac{\dout}{\din}}$, ``shape scaling'' $\max(1,\sqrt{\frac{\dout}{\din}})$, and the RMS-matching factor $0.2\sqrt{\max(\dout,\din)}$ of Kimi/Moonlight~\citep{liu2025muonscalable}, which targets AdamW's per-entry update RMS. In our case (as in the Kimi factor) we use it to match a desired RMS; in our case, that of the weight norm sphere. \autoref{fig:muon-scale-modes} sweeps the matrix LR for each. For plain Muon (\panelleft) every convention clearly beats the AdamW baseline, but the choice of factor still matters noticeably: the unit-RMS-norm factor $\sqrt{\frac{\dout}{\din}}$ and the shape-scaling factor $\max(1,\sqrt{\frac{\dout}{\din}})$ are essentially tied for best, while the RMS-matching factor is noticeably worse. For MuonMD (\panelcenter and \panelright) the shape-up factor $\sqrt{\max\!\big(\frac{\dout}{\din},\, \frac{\din}{\dout}\big)}$ performs best, which we attribute to its matching the RMS of the weight norm under our $\frac{1}{\sqrt{d}}$ initialization; its loss curve stays ahead of the AdamW baseline throughout training. The factor is thus set by the target weight norm and should be adapted whenever that norm changes: for example, the scaled output-projection initialization used in the MoE experiments (\autoref{sec:appendix-setup-moe}) lowers the sphere-norm target of those projections by ${1}/{\sqrt{(2L)}}$ and changes their factor accordingly. Finally, the identification of the initialization norm with $\sqrt{\max(\dout,\din)}$ (and hence the shape-up factor) is exact only when the smaller matrix dimension equals the hidden size $d$; in general the initialization norm is $\frac{1}{\sqrt d}\sqrt{\dout\din}$. This condition generally holds for all projections, except the key/value matrices under GQA, whose output dimension (kv-groups $\times$ head-dim) is smaller than $\din=d$, therefore giving an initialization norm of $\sqrt{\dout}$ rather than $\sqrt{d}$. For these matrices, both the sphere radius and the shape-up factor are larger than intended by $\sqrt{\din/\dout}\approx1.4$ for our half-head KV groups. In our experiments, this discrepancy is negligible.

\begin{figure}[t]
      \centering
      \begin{subfigure}{0.325\textwidth}\centering\includegraphics[width=\textwidth]{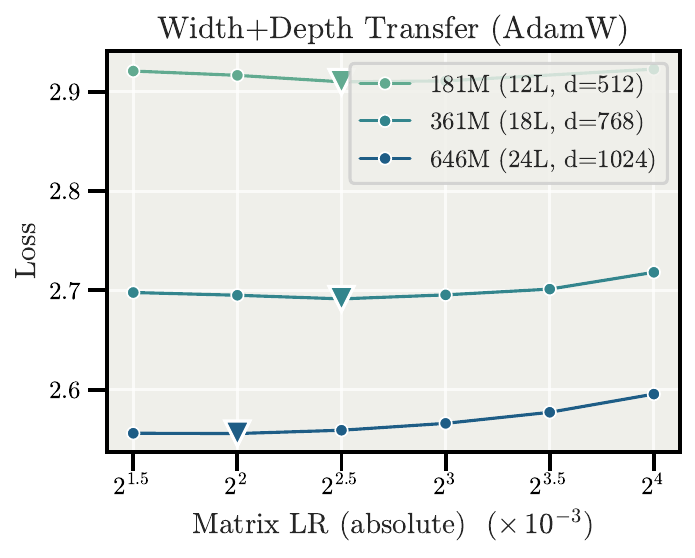}\end{subfigure}\hfill
      \begin{subfigure}{0.325\textwidth}\centering\includegraphics[width=\textwidth]{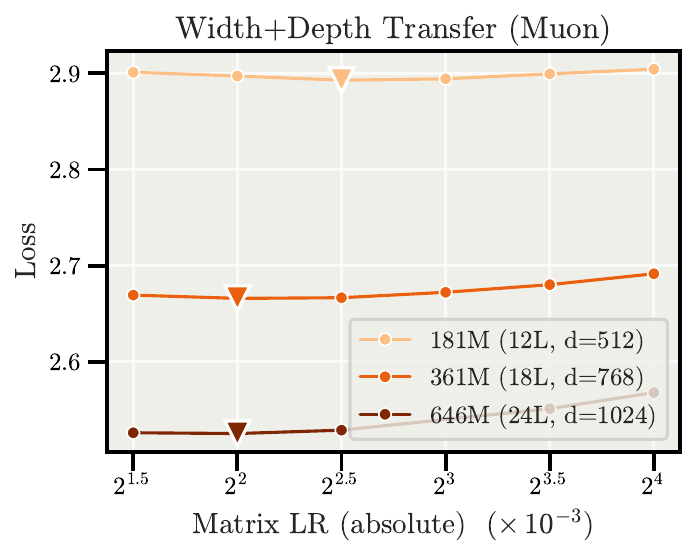}\end{subfigure}\hfill
      \begin{subfigure}{0.325\textwidth}\centering\includegraphics[width=\textwidth]{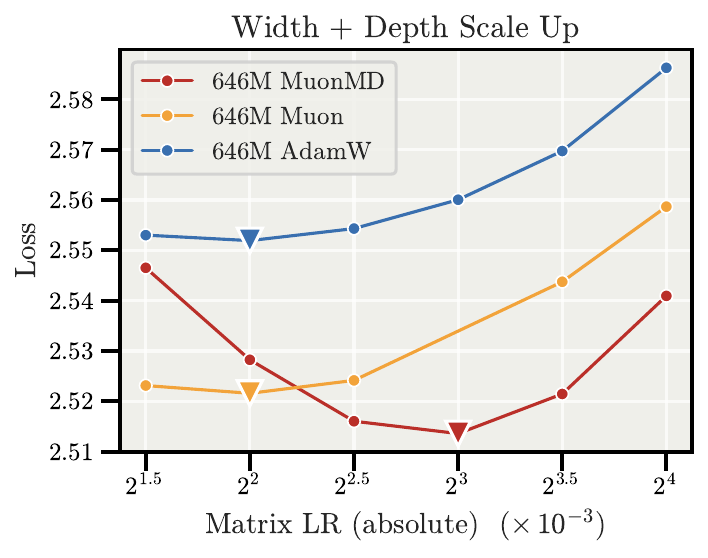}\end{subfigure}
      \caption{\textbf{For reference, the plain AdamW and Muon baselines under joint width-and-depth scaling, where at the largest model MuonMD reaches a lower optimum than both.} Sweeps changing only the matrix LR (no magnitude--direction decoupling). \figleft~AdamW across joint width-and-depth scaling. \figcenter~The same sweep with Muon. \figright~A head-to-head sweep of all three optimizers at the largest joint-scaled model (646M).}
      \label{fig:transfer-baselines}
\end{figure}

\begin{figure}[t]
\centering
\begin{subfigure}{0.325\textwidth}\centering\includegraphics[width=\textwidth]{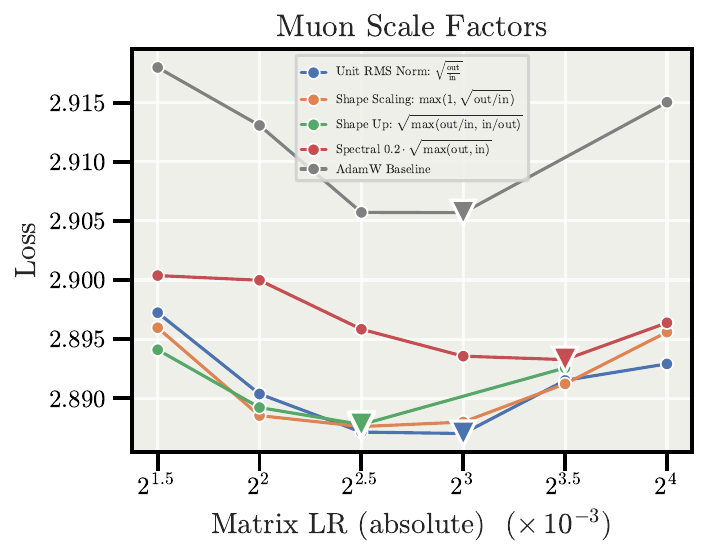}\end{subfigure}\hfill
\begin{subfigure}{0.325\textwidth}\centering\includegraphics[width=\textwidth]{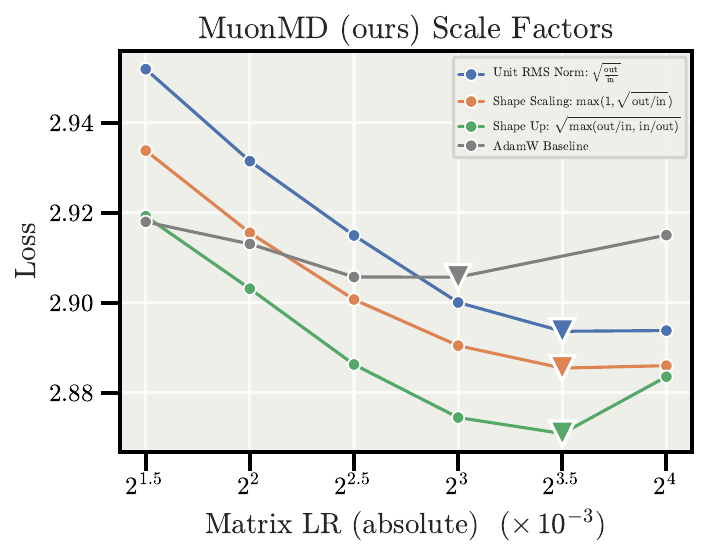}\end{subfigure}\hfill
\begin{subfigure}{0.325\textwidth}\centering\includegraphics[width=\textwidth]{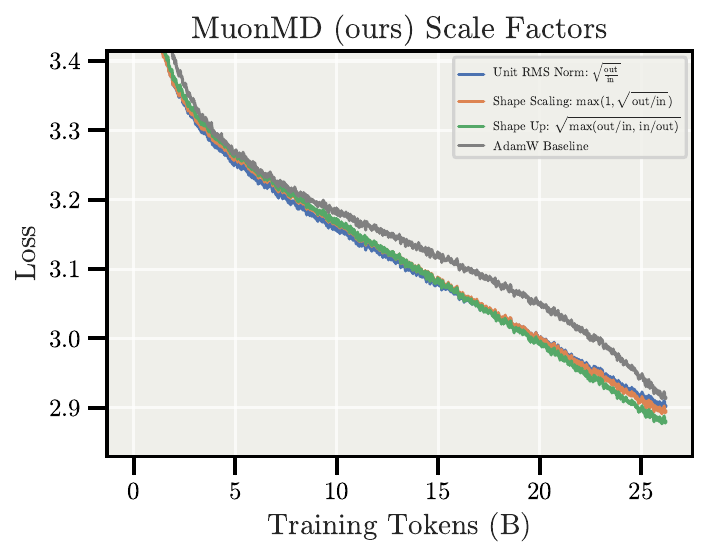}\end{subfigure}
\caption{\textbf{All common Muon scale-factor conventions clearly beat AdamW, but the choice still matters: the unit-RMS-norm $\sqrt{\frac{\dout}{\din}}$ and shape-scaling $\max(1,\sqrt{\frac{\dout}{\din}})$ factors are best and nearly identical, while the RMS-matching factor is noticeably worse.} Sweeps of the matrix LR for each shape-dependent factor that rescales Muon's orthogonalized update, on the 181M model (25B tokens). \figleft~Plain Muon across the scale conventions (with and without warmup); all beat the AdamW baseline, with the unit-RMS-norm and shape-scaling factors tied for best. \figcenter~MuonMD, where we use the shape-up factor $\sqrt{\max\!\big(\frac{\dout}{\din},\, \frac{\din}{\dout}\big)}$ to match the weight-norm RMS. \figright~The corresponding MuonMD loss curves, staying ahead of the AdamW baseline throughout training.}
\label{fig:muon-scale-modes}
\end{figure}

\section{Comparison to nGPT}
\label{sec:appendix-ngpt}

\textbf{Motivation.} Our motivation for holding the weights at a fixed norm is closely related to nGPT~\citep{loshchilov2024ngpt}, which also places its weights on a sphere and drops weight decay, but extends the idea to activations as well. nGPT is, therefore, more than this optimization choice: it bundles the spherical constraint together with a set of architectural changes, making it hard to read off how much of its reported advantage comes from training on the sphere versus from the architecture itself. We therefore aim to compare directly, isolating our optimizer-side recipe from nGPT's architecture, and, where possible, to apply our ideas on top of that architecture to see whether they compose.

\textbf{The nGPT architecture.} In detail, nGPT is a distinct architecture. Beyond constraining the weights to the unit sphere, it (i) replaces every RMSNorm with an $L_2$ normalization, and moves the $L_2$ norms after the attention and MLP blocks and at the end of each layer; (ii) reshapes the residual stream as an interpolation $x + \alpha\,(x' - x)$ toward the normalized block output rather than a plain additive update; (iii) adjusts the attention-logit scaling to compensate for the now $L_2$-normalized queries and keys, as well as scales right before the MLP activation; and (iv) places its 1D learnable vectors (the residual/layer scales, logit scales, etc.) at a reduced base scale, typically $1/\sqrt{d}$. Since each update is relative to that scale, this sharply raises their effective learning rate.

\textbf{Disentangling optimizer and architecture.} We compare on the 181M base model (25B tokens, 50k iterations), matching parameter counts and training budgets and sweeping the matrix LR for every method (\autoref{fig:ngpt}). As proposed, nGPT outperforms our sandwich-norm AdamMD architecture. Applying our optimization ideas \emph{on top of} nGPT's architecture, however, does better still: replacing nGPT's per-vector (row/column) unit-norm projection with our Frobenius constraint already helps, and adding our magnitude gains (AdamMD) helps more, surpassing nGPT. The gap is even larger for Muon --- nGPT's architecture with MuonMD is the best variant overall.

\textbf{Source of nGPT's advantage.} While we have not investigated every single change, we believe that much of nGPT's edge traces back to the (smart) scale trick for its 1D vectors, which our base ablation architecture does not use. Setting the residual/layer scales to $1$ instead of $1/\sqrt{d}$ (grey line, \autoref{fig:ngpt}) removes this effective-LR boost and brings nGPT roughly back down to our base AdamMD, confirming that the trick accounts for a large part of the advantage. We caution, though, that in our experience the very high effective LR induced by ever-smaller scales can become unstable at larger model sizes, so its scalability is unclear; for example, the recent nGPT LR-transfer work~\citep{shigida2026learning} also fixes these scales to be equal across model sizes.

\begin{figure}[t]
\centering
\includegraphics[width=.75\textwidth]{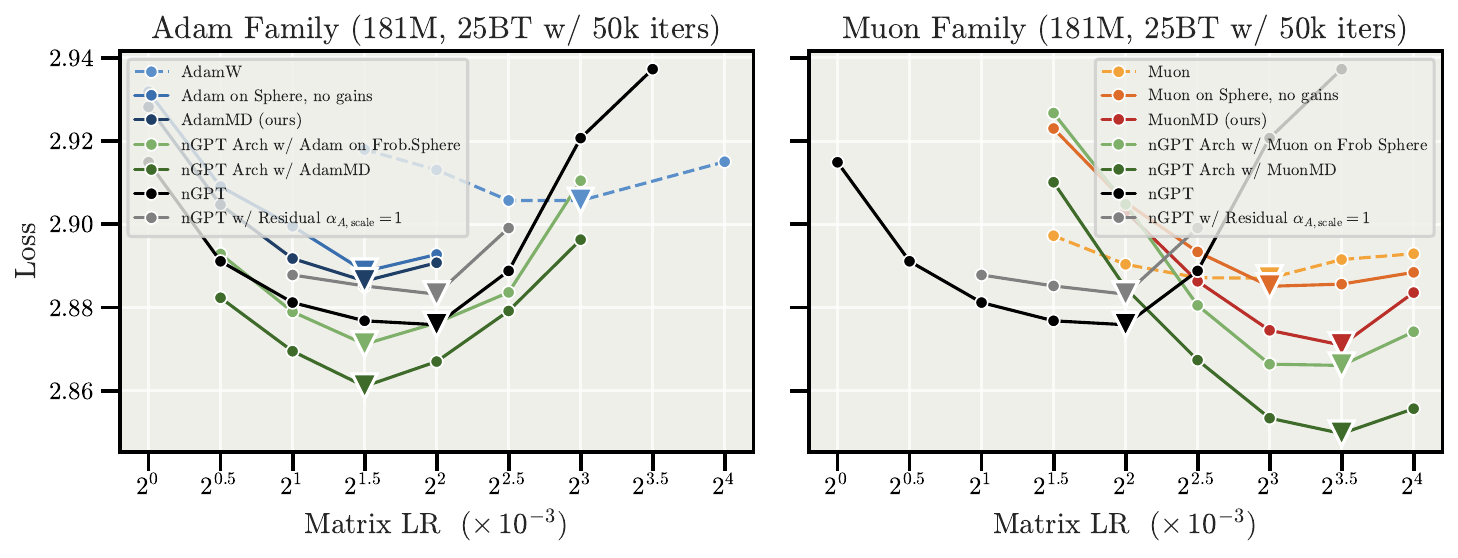}\\[0.6em]
\includegraphics[width=.75\textwidth]{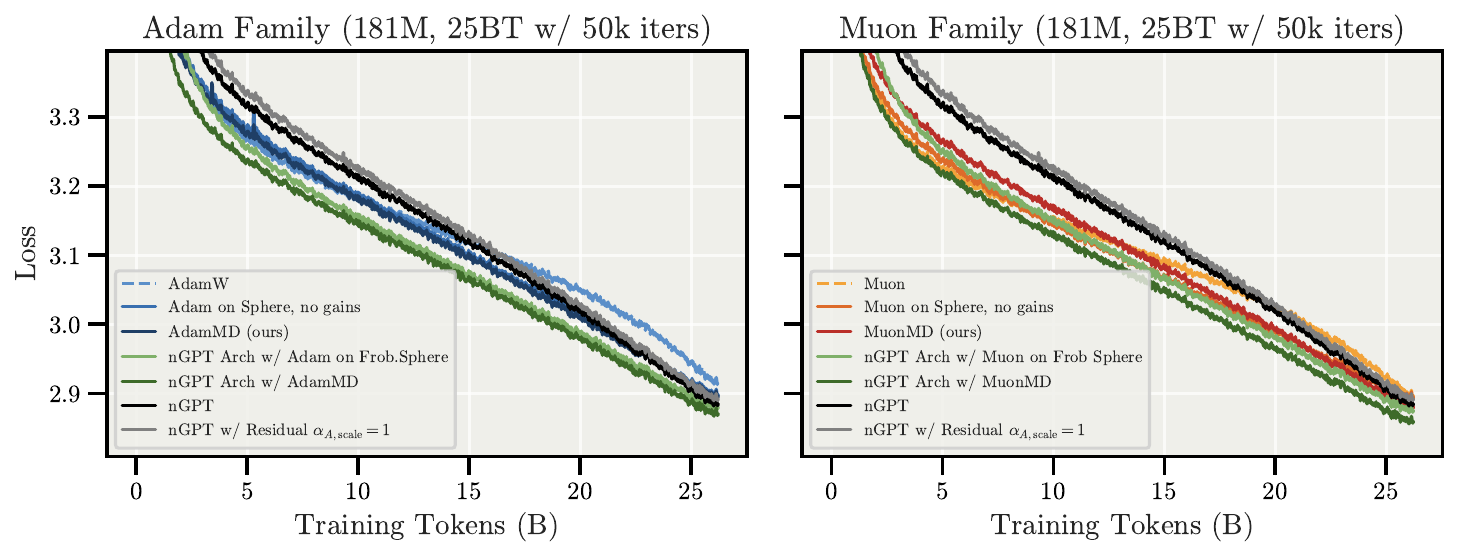}
\caption{\textbf{nGPT is a distinct architecture, not just spherical optimization, and applying our reparameterization on top of it outperforms nGPT as proposed.} Comparison on the 181M model (25B tokens, 50k iters), matched on parameter count and budget. \emph{(Top)}~LR sweeps of the final loss and \emph{(bottom)}~the corresponding loss curves, for the \figleft~Adam family and \figright~Muon family. As proposed by \citet{loshchilov2024ngpt}, nGPT beats our sandwich-norm AdamMD; however, using our Frobenius constraint and magnitude gains on nGPT's architecture (\emph{nGPT Arch w/ AdamMD / MuonMD}) surpasses it, with an even larger margin when using Muon. We believe that much of nGPT's advantage (besides constraining the weights) comes from placing its 1D learnable vectors at a $1/\sqrt{d}$ scale, which raises their effective LR: setting solely the residual scales to $1$ (grey) brings nGPT back down to our base AdamMD.}
\label{fig:ngpt}
\end{figure}

\section{Depth Scaling}
\label{sec:appendix-depth}

We extend the depth-transfer results of \autoref{sec:transfer} by probing the block-output scale $\alpha$ that multiplies each block's output after its RMSNorm. The main text uses $\alpha=\frac{1}{L}$; here we compare it against $\alpha=\frac{1}{\sqrt{2L}}$ across depths from 12 to 30 layers (181M--252M parameters), sweeping the matrix LR at each depth (\autoref{fig:depth-scaling}).

Both choices give good depth transfer: the optimal matrix LR stays roughly fixed across depths rather than drifting with $L$ (\autoref{fig:depth-scaling}, \panelleft). The softer $\frac{1}{\sqrt{2L}}$ scale also gives a small but consistent improvement in final loss at every depth, showing that the precise exponent on $\alpha$ has a noticeable effect on loss even though both choices give transfer without other tricks. A likely explanation shows up in the per-layer activation RMS over training (\panelcenter and \panelright): with $\alpha=\frac{1}{\sqrt{2L}}$ the post-layer activation scale (measured after the MLP residual add) stays controlled around $1$ and roughly uniform across layers, whereas $\alpha=\frac{1}{L}$ lets it drift well below $1$ and spread out across layers in the deep 30-layer model. We leave a fuller study of the optimal $\alpha$ and its interaction with the block-output RMSNorm gains to future work.

\begin{figure}[t]
\centering
\begin{subfigure}{0.325\textwidth}\centering\includegraphics[width=\textwidth]{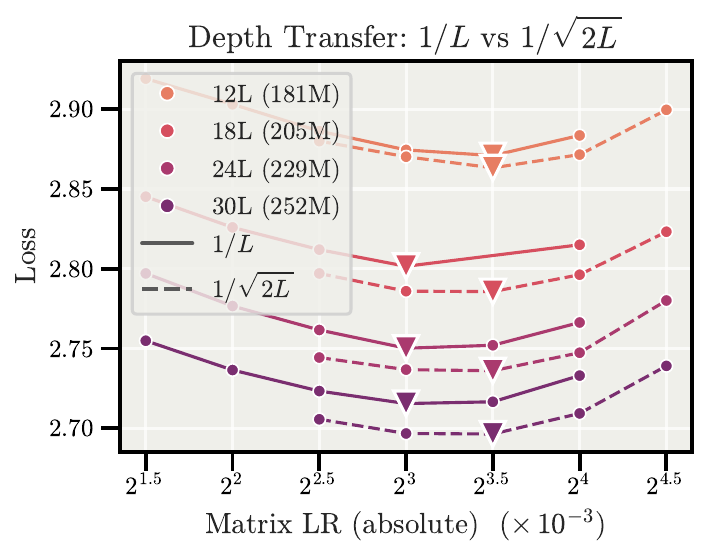}\end{subfigure}\hfill
\begin{subfigure}{0.325\textwidth}\centering\includegraphics[width=\textwidth]{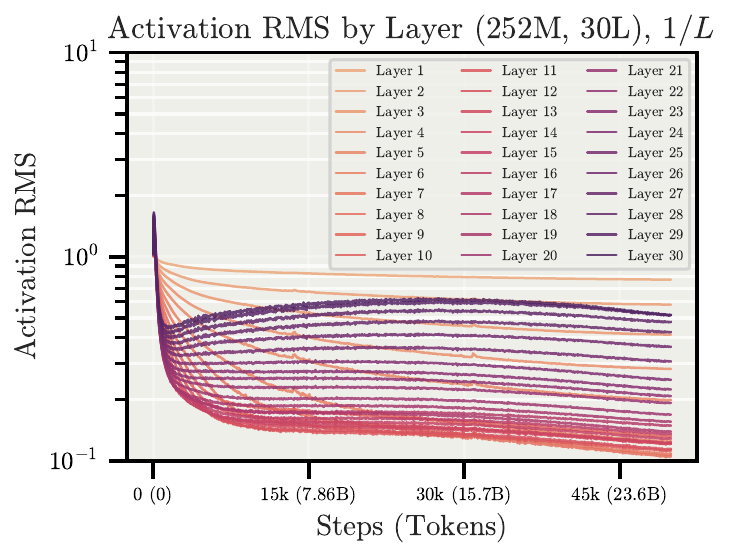}\end{subfigure}\hfill
\begin{subfigure}{0.325\textwidth}\centering\includegraphics[width=\textwidth]{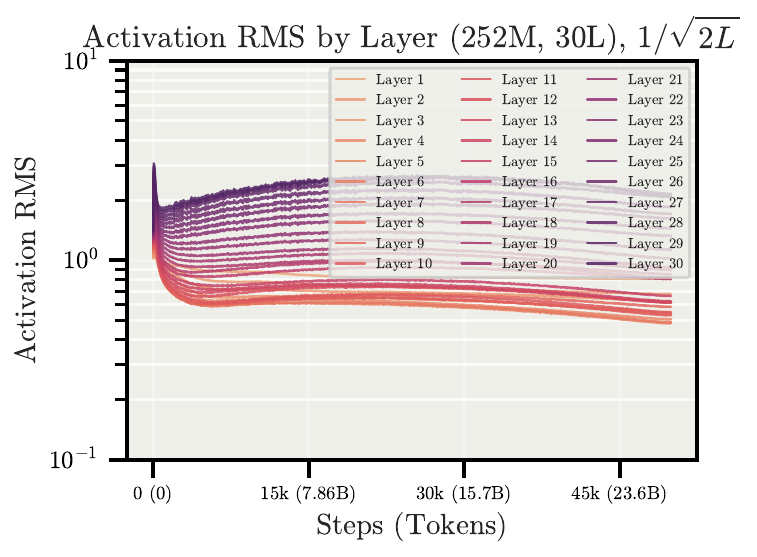}\end{subfigure}
\caption{\textbf{Both block-output scales transfer the matrix LR across depth; the softer $\alpha=\frac{1}{\sqrt{2L}}$ gives a small but consistent loss improvement and keeps per-layer activations better controlled around $1$.} Comparison of $\alpha=\frac{1}{L}$ against $\alpha=\frac{1}{\sqrt{2L}}$. \figleft~LR sweep of the final loss at depths 12--30 (181M--252M parameters), with $\alpha=\frac{1}{L}$ (solid) and $\alpha=\frac{1}{\sqrt{2L}}$ (dashed); the optimal matrix LR stays roughly fixed across depth for both. \figcenter~Per-layer activation RMS over training for the deepest model ($\alpha=\frac{1}{L}$, 252M, 30 layers), drifting well below $1$ and spreading out across layers. \figright~The same for $\alpha=\frac{1}{\sqrt{2L}}$, where the activation RMS stays more clustered around $1$ across layers.}
\label{fig:depth-scaling}
\end{figure}

\section{Implementation, Efficiency, and Throughput}
\label{sec:appendix-efficiency}

\begin{table}[ht]
\centering
\caption{\textbf{Peak stable throughput measured for different optimizers} (in thousands of tokens per second per GPU). The MD variants also show relative slowdown compared to the base optimizer. AdamMD overhead decreases drastically with larger batch sizes and MuonMD overhead remains minimal ($\lesssim 2\%$ for all model sizes).}
\begin{tabular}{lcccc}
\toprule
Params & Adam & AdamMD & Muon & MuonMD \\
\midrule
181M & 185.69 & 176.15 (-5.14\%) & 173.86 & 171.98 (-1.08\%) \\
457M & 117.16 & 111.95 (-4.45\%) & 111.73 & 110.76 (-0.87\%) \\
827M & 74.65 & 70.71 (-5.28\%) & 70.94 & 70.42 (-0.74\%) \\
1.29B & 54.07 & 50.73 (-6.18\%) & 50.82 & 50.55 (-0.53\%) \\
1.54B & 44.08 & 41.47 (-5.92\%) & 41.95 & 41.17 (-1.85\%) \\
1.54B ($2{\times}$GBS) & 44.78 & 43.98 (-1.78\%) & 43.64 & 43.42 (-0.50\%) \\
\bottomrule
\end{tabular}
\label{tab:throughput}
\end{table}

One of the core advantages of MD Decoupling for practical usage lies in the fact that the weights are stored in memory as the fused matrix $W = \diag(\grow)\,\What\,\diag(\gcol)$.
This means that our method adds zero overhead to the architecture during the forward and backward calls.
The additional overhead comes at the optimizer step only, as shown in~\autoref{alg:md-step}, when projecting back $\What$ after the optimizer step, and the additional operations to unfuse, update and fuse the gains $\gamma$.
We note that these operations are all element-wise, and thus represent a small fraction of the total FLOPs of a full training step.

In particular, this overhead remains a fixed fraction of the training step set by the number of tokens processed per optimizer step.
The element-wise cost scales only with the parameter count, while the forward/backward compute scales with both the parameter count and the tokens consumed per step.
Therefore, the relative overhead shrinks as more tokens are processed per optimizer step, e.g., with a larger global batch size or longer sequences.
Within the optimizer step, the fraction shrinks further with hidden dimension in the Muon regime, where the $\mathcal{O}(d^3)$ Newton-Schulz orthogonalization dominates the gain operations.
These operations are moreover memory-bound, which opens the possibility of overlapping the gains-related computation with the Muon NS step.
We leave such optimizations for future work.

We show in \autoref{tab:throughput} the peak stable throughput observed under different optimizers and model sizes for dense models, under equal compute resources.
The architectures used are detailed in~\autoref{tab:dense-arch} (1.54B model size is identical to 1.29B, but using $L=16$ layers instead), and the training was done strictly with data parallelism and measured with two NVIDIA 4xGH200 nodes (DP=8), with distributed optimizer.
In particular, the Adam baseline follows a carefully tuned optimizer state sharding following ZeRO-1~\citep{rajbhandari2020zeromemoryoptimizationstraining}, while Muon and MD variants make use of a layerwise state sharding, resulting in an additional, scale-dependent overhead for the layerwise variants.
The intrinsic gains overhead nonetheless remains small at all scales, as most directly seen in the MuonMD-vs-Muon comparison, where both use layerwise sharding.
Additionally, doubling the global batch size, the 1.54B ($2\times$GBS) configuration, results in a much reduced overhead of $\approx 1.78\%$ in the Adam optimizer.

\section{Higher-Rank Gains}
\label{sec:appendix-higherrank}

\textbf{From rank-1 to rank-$k$.} Our default per-row and per-column gains act on the direction $\What$ through an elementwise multiplicative factor that is effectively \emph{rank-1}: the combined gain $\diag(\grow)\,\What\,\diag(\gcol)$ multiplies entry $(i,j)$ of $\What$ by $\grow[i]\,\gcol[j]$, i.e. by the outer product $\grow\gcol^\top$. This naturally raises the question of whether a rank-$k$ gain matrix --- with finer-grained per-entry control, while still adding far fewer parameters than the weight itself --- performs better. We take first steps in this direction here.

\textbf{Parametrization.} We parameterize the gain matrix $\Gamma\in\R^{\dout\times\din}$ as
\begin{equation}
\Gamma = \mathbf{1} + A B^\top, \qquad A\in\R^{\dout\times k},\ B\in\R^{\din\times k},
\label{eq:higherrank-gain}
\end{equation}
where $\mathbf{1}$ is the all-ones matrix, and the fused weight becomes $W = \Gamma \odot \What$ with $\What$ on the sphere. This is a \emph{direct} parametrization with a $1+\,\cdot$ offset, analogous to the direct gain of \autoref{fig:gain-axis} rather than the softplus map we adopt for the row/column gains. The low-rank factorization $AB^\top$ resembles LoRA~\citep{hu2022lora}, with the key difference that LoRA is \emph{additive} (the low-rank term is added to the weight, $W+AB^\top$), whereas here it is \emph{multiplicative} (it scales the direction elementwise, $\Gamma\odot\What$). We initialize $A$ randomly and $B=0$, so that $\Gamma=\mathbf{1}$ at initialization and the gain leaves the direction untouched at the start of training (as for our scalar/row/column gains, which start at $1$). The factors $A,B$ are updated with Adam at the gain LR, analogously to \autoref{alg:md-step-full}. Here we test $k=4$.

\textbf{Results.} \autoref{fig:higherrank-gains} compares this rank-$k$ ($k=4$) gain against the spherical baseline without gains and against our default per-row/per-column gains, sweeping the matrix LR on the 181M model (25B tokens). The rank-$k$ gain improves over the no-gains baseline, but does not match the softplus row-and-column gain, which remains best. We suspect this gap stems from the parametrization and training dynamics rather than the expressivity. A rank-$k$ gain with $k\geq2$ can represent the rank-1 row-and-column gain exactly, so at $k=4$ it is strictly more expressive, yet still falls short. The likely culprit is that we only try the direct parametrization here (the additive $1+AB^\top$ offset), whereas the row/column gains use the smoother softplus map we found to help (\autoref{fig:gain-axis}). A better parametrization or optimization of the factors $A,B$ may therefore close or reverse this gap without changing what the gain can express. We keep the row-and-column gain as our default and leave a fuller exploration of the rank, the parameterization, and the gain learning rate to future work.

\begin{figure}[t]
\centering
\begin{subfigure}{0.325\textwidth}\centering\includegraphics[width=\textwidth]{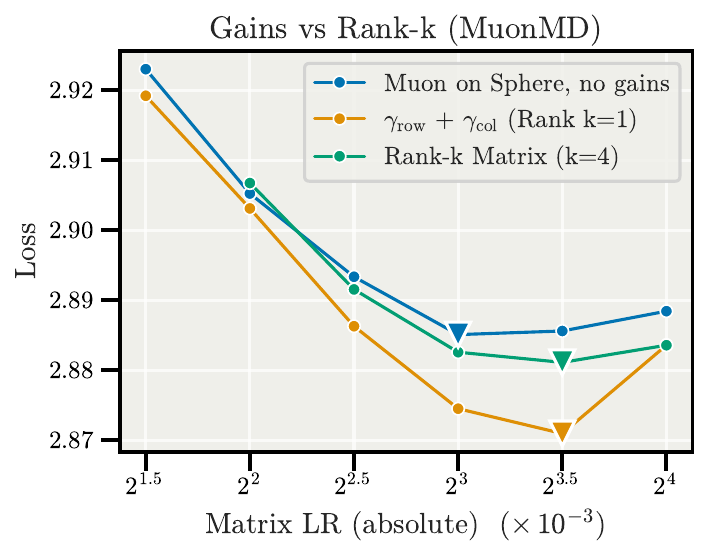}\end{subfigure}\quad
\begin{subfigure}{0.325\textwidth}\centering\includegraphics[width=\textwidth]{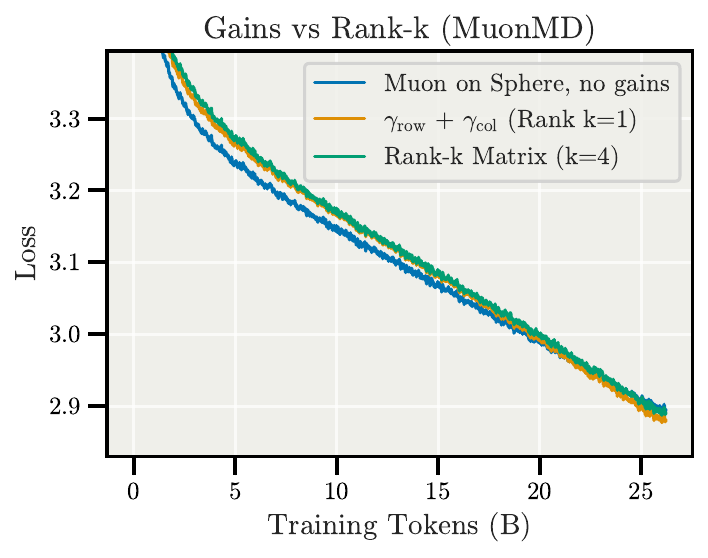}\end{subfigure}
\caption{\textbf{A higher-rank gain improves over no gains but does not beat the simpler row-and-column gain, which remains our default.} MuonMD on the 181M model (25B tokens), comparing the spherical baseline without gains, our default per-row/per-column gain $\grow+\gcol$, and a rank-$k$ gain matrix $\Gamma=\mathbf{1}+AB^\top$ with $k=4$. \figleft~LR sweep of the final loss. \figright~The corresponding loss curves.}
\label{fig:higherrank-gains}
\end{figure}

\section{Gain Dynamics}
\label{sec:appendix-gain-dynamics}

The gains $\grow,\gcol$ add a learnable magnitude on top of the fixed-norm direction, and the main text settles which gain mode and parameterization to use (\autoref{fig:gain-axis}). Here we take a brief, descriptive look at how these gains behave over the course of training. \autoref{fig:gain-dynamics} tracks three representative projections at layer 6 of the 181M dense model (the attention QKV projection, the MLP up-and-gate projection (\texttt{linear\_fc1}), and the MLP output projection (\texttt{linear\_fc2})) for the four parameterizations of \autoref{fig:gain-axis} (the direct update with and without a $10^{-5}$ floor, the exponential map, and softplus). We plot the raw-gain norm $\lVert\widehat{\gamma}\rVert_2$ and the minimum and maximum effective gain $\varphi(\widehat{\gamma})$ over training for the row and column gains.

\textbf{The gains are actively used.} In every projection, the raw-gain norm grows throughout training and the effective gains spread out over more than an order of magnitude (here from roughly $0.1$ up to $5$--$7$), so the model moves the magnitudes well away from their initialization of $1$ rather than leaving them put. This holds both for \texttt{linear\_qkv}, which is following the pre-norm and feeds into QK-norm, \texttt{linear\_fc1}, whose output feeds the SwiGLU nonlinearity with \emph{no} normalization layer afterward, and \texttt{linear\_fc2}, which writes into the residual stream and is followed by the post-MLP Sandwich Norm. This is evidence that the gains add control beyond what normalization already provides.

\textbf{Parameterization is benign.} The parameterizations differ just as the shape of each map $\varphi$ would suggest: the exponential map grows the widest range of effective gains, while softplus stays the most contained, and the unfloored direct update even lets some gains cross zero and flip sign (its minimum settles around $-1.5$). Perhaps surprisingly, as the main text shows (\autoref{fig:gain-axis}), this leaves the loss and the overall dynamics essentially unchanged. We make no claim to a mechanistic understanding of \emph{why} per-row and per-column scales help or what role the model assigns to the large and small gains, and leave these interpretability questions to future work.

\section{Extended Related Work}
\label{sec:appendix-relwork}

This section expands the \emph{broader context} paragraph of the related work in \autoref{sec:related}.

\textbf{Sphere and manifold constraints for LLM training.} A connected line of work constrains the weights (and sometimes updates) to a sphere or matrix manifold during pretraining, differing mainly in \emph{which} norm is fixed and whether magnitude is added back. On a per-vector or Frobenius sphere: nGPT~\citep{loshchilov2024ngpt} normalizes every weight's rows/columns (depending on up/down projection) as well as activations to a fixed $L_2$ norm of 1 (different to RMSNorms), alongside further architectural changes we dissect and compare against in \autoref{sec:appendix-ngpt}; anGPT~\citep{franke2025learning} relaxes this to an approximate norm constraint, and ~\citet{fishman2026normalized} find how nGPT enables low-precision training. The same activations-on-the-sphere idea has also been used for generation, where \citet{deschenaux2026hyperspherical} learn a latent flow that produces language by transporting token representations along a velocity field on the hypersphere. Nemotron-Flash~\citep{fu2025nemotron} keeps the nGPT unit-norm projections without the other architectural changes; and Mano~\citep{gu2026mano} projects the momentum onto the tangent space of a rotational oblique manifold (alternating unit-norm columns and rows). On the \emph{spectral} norm instead: SSO~\citep{xie2026controlled} constrains both weights and updates, Modular Manifolds~\citep{bernstein2025modular} co-designs the optimizer with a Stiefel constraint, and Enforced Lipschitz Constants~\citep{newhouse2025training} bounds the operator norm throughout training. These manifold constraints are connected to a broader literature on optimization over the orthogonal/Stiefel manifold, e.g.\ retraction-free methods that stay on the manifold without an expensive projection step~\citep{ablin2022fast}; the same polar-decomposition view of separating an orthonormal direction from a scale also underlies parameter-efficient fine-tuning, where PoLAR~\citep{lion2026polar} constrains each low-rank adapter's factors to the Stiefel manifold and learns a separate core scale. In our work, we instead fix the softer Frobenius norm and add learnable magnitudes inside the optimizer for general pretraining, independent of how the update step is obtained. Relatedly, width scaling under operator norms~\citep{xu2026width} obtains the same LR transfer from an operator-norm view, while Target Variance Rescaling~\citep{owen2025variance} periodically rescales to a target variance rather than a norm.

\textbf{Controlling the relative update.} Previous work aims to control the update size \emph{relative} to the weight, without an explicit magnitude/direction split. The underlying dynamics were first studied for scale-invariant weights under normalization, where weight decay and the norm together set an ``effective learning rate''~\citep{vanlaarhoven2017l2,hoffer2018norm,arora2019theoretical,li2020exponential,wan2021spherical,kodryan2022training,kosson2026balanced}. Prior work argued that controlling the relative weight update is the main effect of weight decay, creating Rotational Optimizer Variants (RVs)~\citep{kosson2023rotational} and LionAR~\citep{kosson2024analyzing} that achieve the same effect by fixing the weight norm and scaling the update norm to be proportional on average. Nero~\citep{liu2021learning} was an earlier optimizer that used a similar mechanism of constraining the norms and controlling the update norm of each neuron without specifically relating it to weight decay. LARS~\citep{you2017large} and LAMB~\citep{you2019large}, and variants of AdaFactor~\citep{zhai2021scaling} scale the update of each layer to be proportional to the weight norm without explicitly constraining it. In RL, Normalize-and-Project~\citep{lyle2024normalization} periodically projects weights back to their initial per-layer norm to keep the effective LR constant, and SimbaV2~\citep{lee2025hyperspherical} normalizes weights and features onto a hypersphere to scale up RL agents --- like our sphere constraint, but motivated by plasticity and stability rather than transfer. For diffusion models, EDM2~\citep{karras2023analyzing} combines weight projections and normalization layers to keep relative updates from decaying over time and balance their size between layers. AdamP/SGDP~\citep{heo2021adamp} removed the radial component of the update stemming from momentum to slow down the magnitude growth without explicitly constraining the norm.

\textbf{Hyperparameter transfer and warmup.} A parallel line of work transfers hyperparameters across scale rather than retuning them. The maximal-update parametrization ($\mu$P)~\citep{yang2022tensor} and its depth extension~\citep{yang2023tensor} transfer the optimal LR across width and depth, with later analyses mapping how the right exponents depend on the optimizer and parametrization~\citep{everett2024scaling,dey2025completep,mlodozeniec2025completed,ren2026rethinking}; we instead obtain width transfer directly from the sphere constraint. Related to our motivation, $\nu$GPT~\citep{shigida2026learning} restores LR transfer for the normalized-transformer (nGPT) architecture by combining $\mu$P with alignment exponents. A related thread investigates how \emph{weight decay} and batch size --- not just the LR --- influences learning and how it should scale, and how whole loss curves collapse onto a universal trajectory under the right recipe~\citep{andriushchenko2023weight,wang2024adamw,bergsma2025power,bergsma2025collapse}; MD Decoupling instead removes weight-decay tuning altogether. Warmup is another near-universal ingredient we are able to drop: it has been explained both as preventing instability in the deeper layers~\citep{gotmare2019closer} and as a variance-reduction device for adaptive optimizers in their early steps~\citep{goyal2017accurate,liu2020variance,xiong2020layer}, neither of which generally arises once the updates are normalized and the weights stay on the sphere.

\textbf{Classic reparameterization and normalization.} The idea of separating a weight's magnitude from its direction goes back to before LLMs. Weight Normalization~\citep{salimans2016weight} reparameterizes each weight as $w = (g/\lVert v\rVert)\,v$ --- a learnable scalar magnitude $g$ times a direction. This is the most direct classic ancestor of our gains, though without a fixed-norm constraint or a separate LR for the direction. Decoupled Networks~\citep{liu2018decoupled} factor the neuron's inner product into a magnitude function times an angular function of the angle between weight and input. Weight Standardization~\citep{qiao2019micro} and its use in BiT~\citep{kolesnikov2020big} standardize the weights feeding each output channel to zero mean and unit variance to smooth the loss landscape, and Spectral Normalization~\citep{miyato2018spectral} divides each matrix by its top singular value to bound the Lipschitz constant in GANs. Hyperspherical units, e.g. AKOrN~\citep{miyato2025artificial}, keep their state vectors on a sphere by construction. These methods normalize or reparameterize weights for conditioning, stability, or robustness. Our contribution is to put the direction on a \emph{fixed} sphere with a normalized update \textbf{and} learn per-row/per-column magnitudes at their own rate, specifically to remove the magnitude--direction interference and improve training performance.

\begin{figure}[p]
\centering
\begin{subfigure}{\textwidth}\centering\includegraphics[width=0.72\textwidth]{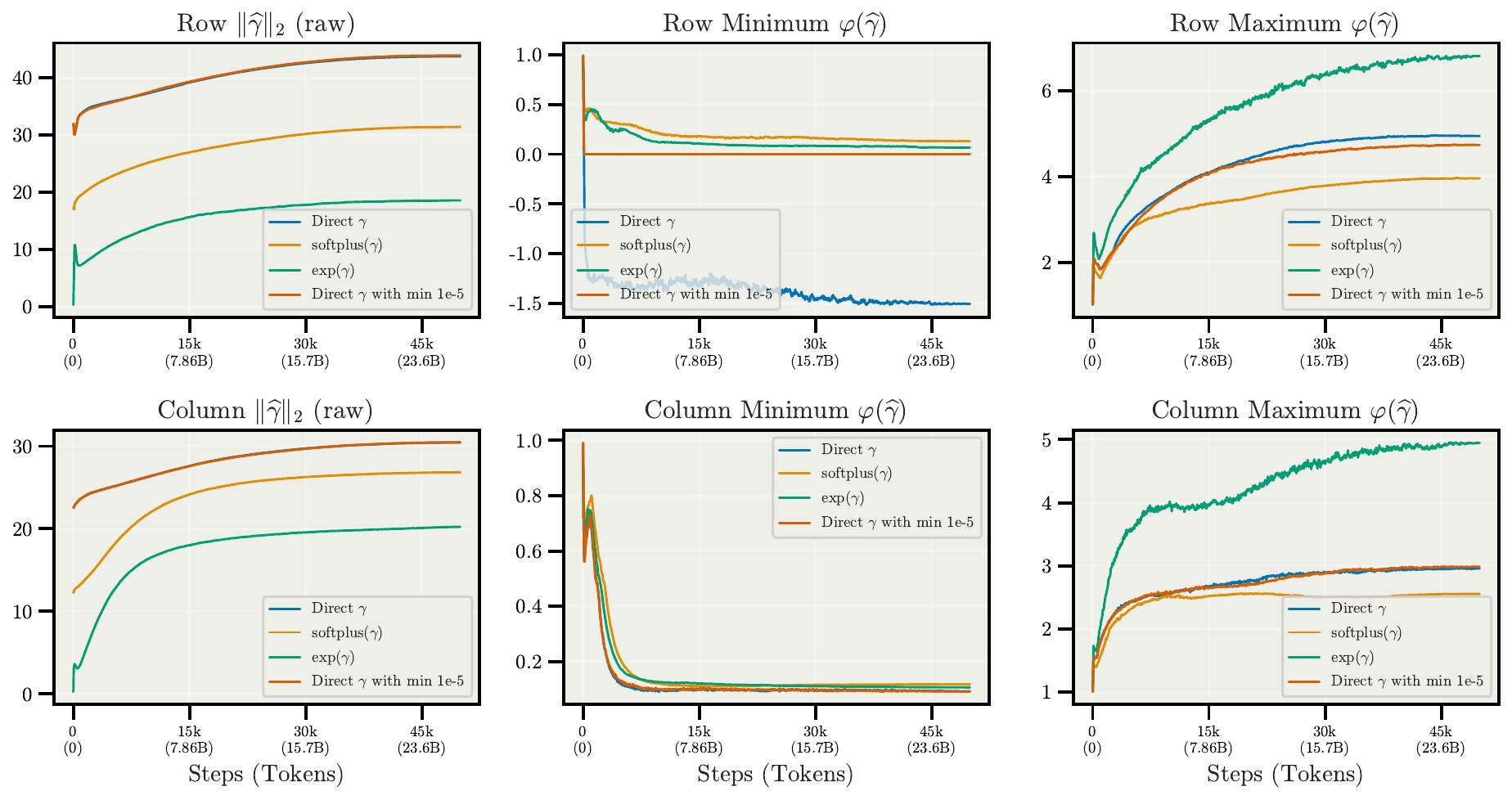}\caption{\footnotesize Attention QKV projection.}\end{subfigure}
\begin{subfigure}{\textwidth}\centering\includegraphics[width=0.72\textwidth]{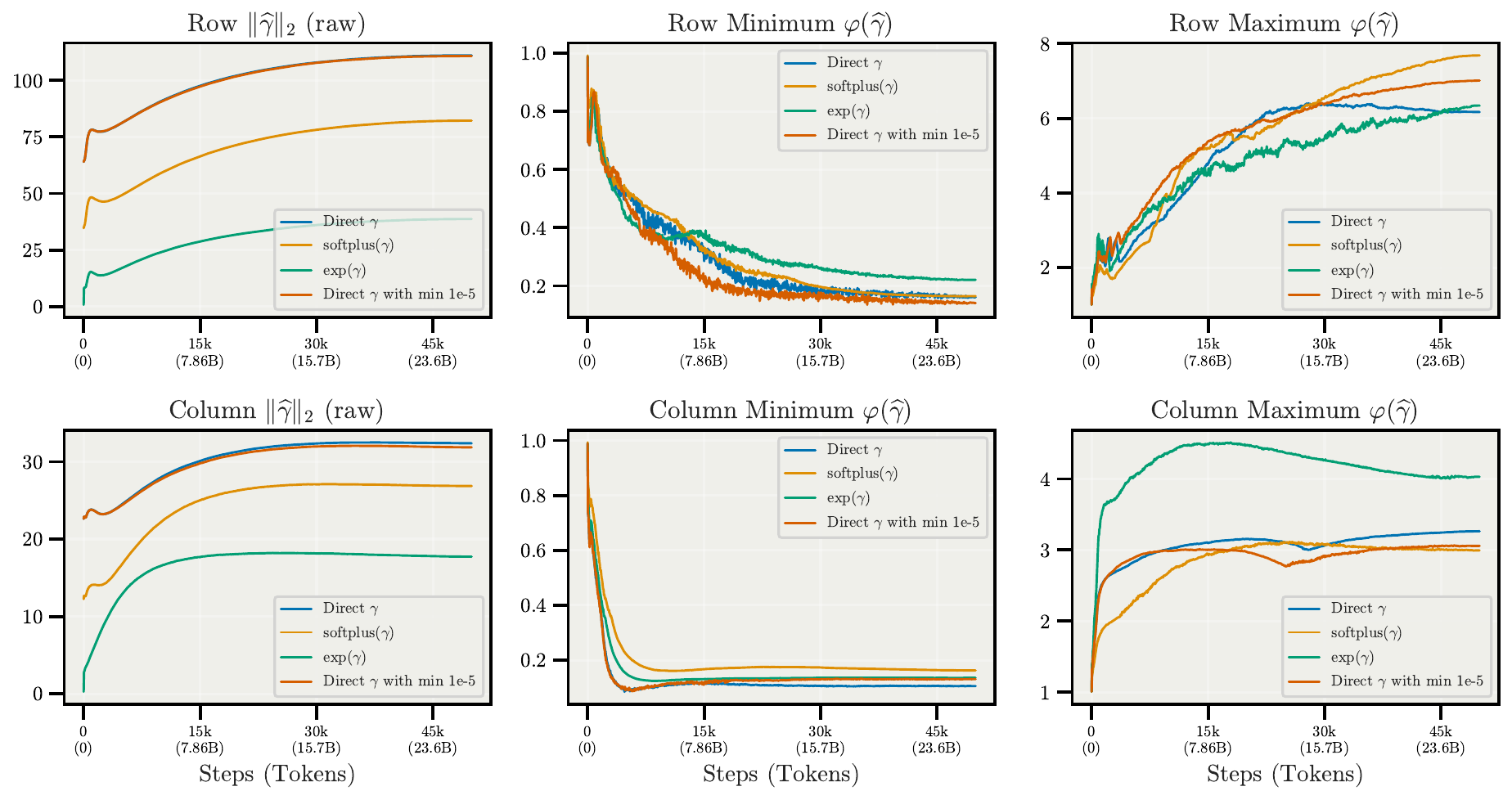}\caption{\footnotesize MLP up-and-gate projection (\texttt{linear\_fc1}); no normalization layer downstream.}\end{subfigure}
\begin{subfigure}{\textwidth}\centering\includegraphics[width=0.72\textwidth]{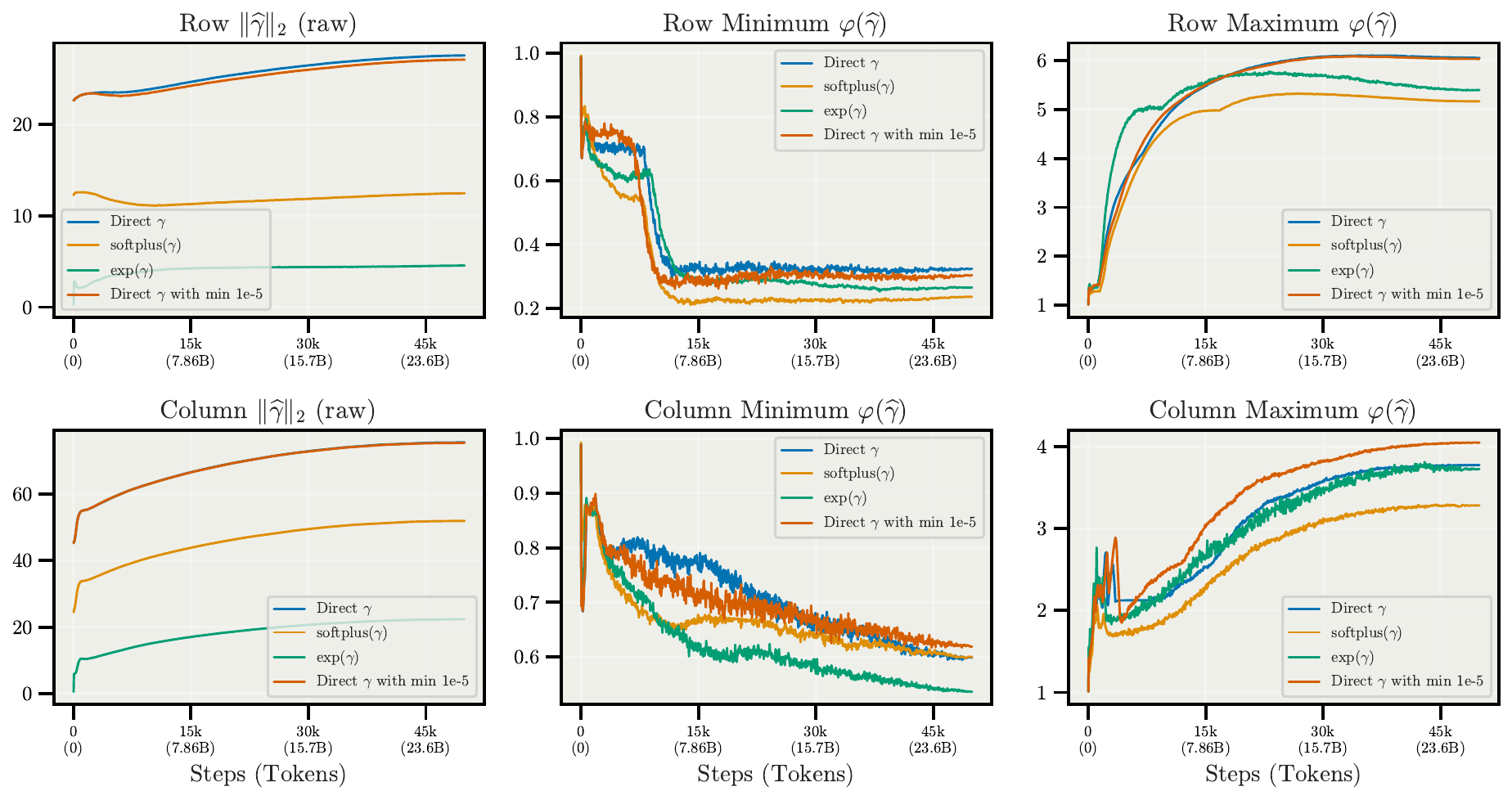}\caption{\footnotesize MLP output projection (\texttt{linear\_fc2}); followed by the post-MLP Sandwich Norm.}\end{subfigure}
\caption{\textbf{Across all three projections, the learned gains spread over more than an order of magnitude during training.} Gain dynamics at layer 6 of the 181M dense model, for the four parameterizations of \autoref{fig:gain-axis}. In each panel, \emph{(top)}~row and \emph{(bottom)}~column gains; \figleft~The raw-gain norm $\lVert\widehat{\gamma}\rVert_2$, \figcenter~The minimum effective gain $\varphi(\widehat{\gamma})$ (over all dimensions), and \figright~The maximum effective gain (over all dimensions).}
\label{fig:gain-dynamics}
\end{figure}

\end{document}

%% file: math_commands.tex
\usepackage{amsmath,amsfonts,bm}

\usepackage{xspace}
\newcommand{\figleft}{{\em (Left)}\xspace}
\newcommand{\figcenter}{{\em (Center)}\xspace}
\newcommand{\figright}{{\em (Right)}\xspace}

\newcommand{\panelleft}{left\xspace}
\newcommand{\panelcenter}{center\xspace}
\newcommand{\panelright}{right\xspace}

\def\eqref#1{equation~\ref{#1}}

\def\1{\bm{1}}

\DeclareMathAlphabet{\mathsfit}{\encodingdefault}{\sfdefault}{m}{sl}
\SetMathAlphabet{\mathsfit}{bold}{\encodingdefault}{\sfdefault}{bx}{n}

\newcommand{\R}{\mathbb{R}}